\PassOptionsToPackage{numbers,sort&compress}{natbib}
\PassOptionsToPackage{main=english,russian}{babel} 
\documentclass{article}
\usepackage{float}
\usepackage{listings}

\usepackage[preprint]{style}
\makeatletter
\providecommand{\@noticestring}{}
\makeatother

\usepackage{etoolbox}
\AtBeginDocument{%
  \ifdefined\linenumbers
    \patchcmd{\equation}{\tagform@}{\linenomath\tagform@}{}{}%
    \patchcmd{\endequation}{\endlinenomath}{}{}{}%
  \fi
}

\usepackage{listings}
\usepackage{xcolor}
\definecolor{darkred}{RGB}{139,0,0}
\definecolor{darkgreen}{RGB}{0,100,0}
\usepackage{bm}
\usepackage{lmodern}

\usepackage[utf8]{inputenc} 
\usepackage[T1]{fontenc}    
\usepackage{hyperref}       
\usepackage{graphicx}       
\usepackage{url}            
\usepackage{booktabs}       
\usepackage{amsfonts}       
\usepackage{nicefrac}       
\usepackage{microtype}      
\usepackage{xcolor}         
\usepackage{amsmath}
\usepackage{wrapfig} 
\usepackage[export]{adjustbox}  
\usepackage{comment}
\usepackage{algorithm}
\usepackage{algpseudocode}
\DeclareUnicodeCharacter{2061}{} 
\usepackage{xcolor}
\usepackage{listings}
\usepackage{tcolorbox}
\tcbuselibrary{listings}      

\let\origttdefault\ttdefault
\usepackage{sourcecodepro} 
\renewcommand{\ttdefault}{\origttdefault} 

\definecolor{vs-bg}        {HTML}{FDFDFD}   
\definecolor{vs-grayframe} {HTML}{D0D0D0}   

\definecolor{vs-keyword}   {HTML}{0000FF}   
\definecolor{vs-string}    {HTML}{A31515}   
\definecolor{vs-comment}   {HTML}{008000}   

\definecolor{funcbrown}{HTML}{795E26}   
\definecolor{varteal}  {HTML}{267F99}   
\definecolor{comment}  {HTML}{6A9955}   
\definecolor{arrowmag} {HTML}{000000}   
\definecolor{boolpurple}{HTML}{AF00DB}  
\definecolor{vscallblue}{HTML}{267F99}
\definecolor{vs-identblue}{HTML}{000C60} 

\newcommand{\rarrow}{\textcolor{arrowmag}{${\rightarrow}$}}

\usepackage{bm}
\renewcommand{\rarrow}{\textcolor{arrowmag}{$\bm{\rightarrow}$}}
\lstdefinestyle{pythonish}{
  language=Python,
  basicstyle=\ttfamily\footnotesize\color{black},
  backgroundcolor=\color{vs-bg},
  columns=fullflexible, tabsize=2, keepspaces, showstringspaces=false, breaklines,
  keywordstyle=\color{vs-keyword}\bfseries,
  stringstyle=\color{vs-string},
  commentstyle=\color{vs-comment}\itshape,
  morekeywords=[2]{Linear,BatchNorm1d,ReLU,lmKANLayer},
  morekeywords=[2]{Sequential,RandomCrop,RandomHorizontalFlip,ColorJitter,RandAugment,RandomErasing,Normalize},
  keywordstyle=[2]\color{vscallblue},
  morekeywords=[3]{True,False},
  keywordstyle=[3]\color{vs-keyword},
  morekeywords=[4]{input_dim,output_dim,hidden_dim,affine},
  morekeywords=[4]{padding,p,scale,ratio,MEAN,STD},
  keywordstyle=[4]\color{vs-identblue},
  literate={->}{{\rarrow}}2
}

\newtcblisting{archbox}[1]{%
  colback=vs-bg,
  colframe=vs-grayframe,
  arc=1.5mm,
  title={\textbf{#1}},
  fonttitle=\sffamily\bfseries\color{black}, 
  listing engine=listings,
  listing only,
  listing options={style=pythonish},
  left=2pt,right=2pt,top=-5pt,bottom=-5pt
}

\algrenewcommand\algorithmicrequire{\textbf{Input:}}
\algrenewcommand\algorithmicensure{\textbf{Output:}}
\algrenewcommand\algorithmiccomment[1]{\hfill\(\triangleright\) #1}

\lstdefinestyle{pythonish-aug}{
  style=pythonish,
  morekeywords   =[2]{Sequential,RandomCrop,RandomHorizontalFlip,ColorJitter,
                     RandAugment,RandomErasing,Normalize},
  morekeywords   =[4]{padding,p,scale,ratio,MEAN,STD},
}

\newtcblisting{archboxaug}[1]{%
  colback=vs-bg, colframe=vs-grayframe, arc=1.5mm,
  title={\textbf{#1}}, fonttitle=\sffamily\bfseries\color{black},
  listing engine=listings, listing only,
  listing options={style=pythonish-aug},
  left=2pt,right=2pt,top=-5pt,bottom=-5pt
}

\lstdefinestyle{pythonish-cnn}{
  style=pythonish,
  morekeywords=[2]{Conv2D,FullyConnected},
  morekeywords=[4]{width,kernel_size,stride},
}

\newtcblisting{archboxcnn}[1]{%
  colback=vs-bg, colframe=vs-grayframe, arc=1.5mm,
  title={\textbf{#1}}, fonttitle=\sffamily\bfseries\color{black},
  listing engine=listings, listing only,
  listing options={style=pythonish-cnn},
  left=2pt,right=2pt,top=-5pt,bottom=-5pt
}

\lstdefinestyle{pythonish-cnn-extended}{
  style=pythonish-cnn,
  morekeywords=[4]{base_width},
}

\newtcblisting{archboxcnnX}[1]{%
  colback=vs-bg, colframe=vs-grayframe, arc=1.5mm,
  title={\textbf{#1}}, fonttitle=\sffamily\bfseries\color{black},
  listing engine=listings, listing only,
  listing options={style=pythonish-cnn-extended},
  left=2pt,right=2pt,top=-5pt,bottom=-5pt
}

\hypersetup{colorlinks=true, urlcolor=black, citecolor=black, linkcolor=black}
\title{Lookup multivariate Kolmogorov-Arnold Networks}

\author{
Sergey Pozdnyakov$^1$ \\
  \texttt{sergey.pozdnyakov@epfl.ch} \\
  \And
  Philippe Schwaller$^{1,2}$ \\
  \texttt{philippe.schwaller@epfl.ch} \\
  \and \\ [-1.5em]
$^1$ École Polytechnique Fédérale de Lausanne (EPFL) \\
$^2$ National Centre of Competence in Research (NCCR) Catalysis
}
\begin{document}
\maketitle
\begin{abstract}
High-dimensional linear mappings, or linear layers, dominate both the parameter count and the computational cost of most modern deep-learning models. We introduce a general-purpose drop-in replacement, lookup multivariate Kolmogorov-Arnold Networks (lmKANs), which deliver a substantially better trade-off between capacity and inference cost. Our construction expresses a general high-dimensional mapping through trainable low-dimensional multivariate functions. These functions can carry dozens or hundreds of trainable parameters each, and yet it takes only a few multiplications to compute them because they are implemented as spline lookup tables. Empirically, lmKANs reduce inference FLOPs by up to 6.0× while matching the flexibility of MLPs in general high-dimensional function approximation. In another feedforward fully connected benchmark, on the tabular-like dataset of randomly displaced methane configurations, lmKANs enable more than 10× higher H100 throughput at equal accuracy. Within frameworks of Convolutional Neural Networks, lmKAN-based CNNs cut inference FLOPs at matched accuracy by 1.6–2.1× and by 1.7× on the CIFAR-10 and ImageNet-1k datasets, respectively. Our code, including dedicated CUDA kernels, is available online at \url{https://github.com/schwallergroup/lmkan}. 
\end{abstract}

\begin{figure}[H]
    \centering
    \includegraphics[width=0.71\linewidth]{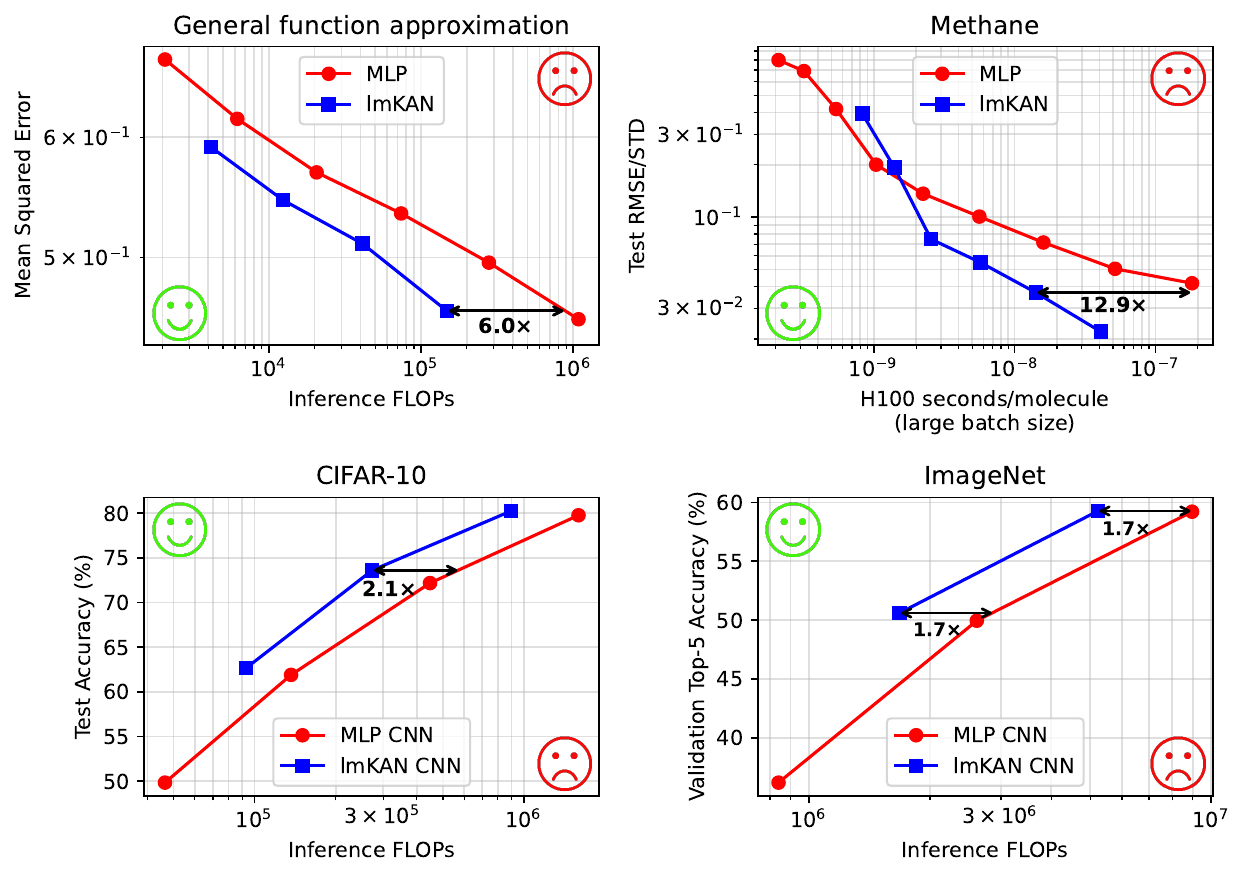}
    \caption{Performance summary. See Sec.~\ref{sec:experiments} for details.}
    \label{fig:combined_plot}
\end{figure}

\section{Introduction}
\label{sec:introduction}

With a sufficient amount of training data, the capabilities of deep-learning models systematically improve with the number of trainable parameters~\citep{zhai2022scaling, kaplan2020scaling}. However, deploying very large models is challenging because of the associated inference cost.

In most models, high-dimensional linear mappings dominate both the parameter count and the computational cost. Standard multilayer perceptrons (MLPs) alternate linear layers with activations, and sometimes with a few other layers~\citep{ioffe2015batch,hinton2012improving}. If $N$ is the width of a layer, the parameter count and inference cost of these linear mappings scale as $\mathcal{O}(N^{2})$, whereas most other layers scale only as $\mathcal{O}(N)$. The same observation holds for many other architectures. Transformers~\citep{vaswani2017attention}, when applied to very long sequences, are one of the few notable exceptions because the cost of attention grows quadratically with the number of tokens. Even in that case, however, the cost of the linear mappings remains substantial, not to mention the potential use of fast approximations of attention~\citep{choromanski2020rethinking}.

The computational cost of a linear layer is proportional to the number of its parameters: at inference, each parameter induces one multiplication per input, where the input is a whole input object in the case of MLPs, a token in the case of Recurrent Neural Networks~\citep{elman1990finding} and Transformers~\citep{vaswani2017attention}, a node or an edge in the case of Graph Neural Networks~\citep{zhou2020graph}, a patch of an image in the case of Convolutional Neural Networks~\citep{lecun2002gradient, krizhevsky2012imagenet}, and similarly for other architectures.  

Spline lookup tables make it possible to do better than that. Consider, for example, a one-dimensional piecewise-linear function $f(x)$ on the interval from $0$ to $1$ with a uniform grid. On each interval, it is given as $f(x) = a[i] * x + b[i]$, where $i$ is the interval index. With $G$ intervals, the function has $2G$ trainable parameters, out of which $G+1$ are independent once continuity at the internal grid points is enforced. Yet the computational cost of evaluating such a function at any given point is $\mathcal{O}(1)$, not depending on $G$. The computational pipeline involves first determining the current grid interval as $i = \lfloor x * G\rfloor$, and then evaluating only one linear function as $f(x) = a[i] * x + b[i]$. 


Kolmogorov-Arnold Networks (KANs)~\citep{liu2024kan}, designed as a general alternative to MLPs, are natural hosts for spline lookup tables as they construct a general high-dimensional mapping through a collection of trainable univariate functions. 

The main contributions of this work are the following:
\begin{itemize}
    \item We propose lookup multivariate Kolmogorov-Arnold Networks (lmKANs) that are built upon multivariate low-dimensional functions instead of the univariate ones that standard KANs employ. We empirically compare the 2D version of lmKANs with 1D FastKAN~\citep{li2024kolmogorovarnold} and find that lmKANs are more accurate and easier to train. 
    \item We implement the inner functions as spline lookup tables. Ignoring a non-asymptotic $\mathcal{O}(N)$ term, the required inference FLOPs are only $2\times$ those of a linear layer of the same shape, while the number of trainable parameters can be dozens or hundreds of times higher.
    \item We provide custom CUDA kernels that enable efficient inference of lmKANs on modern GPUs. When using the $8\times8$ tile size, on the H100 GPU, our implementation enables up to $\sim88\times$ faster inference than a linear layer with the same number of trainable parameters.
    \item We empirically compare lmKANs and MLPs across diverse datasets, scales, and backbones, using varied experimental setups to obtain a comprehensive view of performance. Across these conditions, lmKANs are consistently Pareto-optimal with respect to inference FLOPs. The performance of lmKANs is summarized in Fig.~\ref{fig:combined_plot}.
\end{itemize}

The proposed lmKAN layers can serve as a drop-in replacement for high-dimensional linear mappings across a wide range of deep-learning architectures.

\section{Related work}
\label{sec:related_work}
Kolmogorov-Arnold Representation Theorem (KART)~\citep{kolmogorov1961representation, arnold2009functions} states that a continuous function $f : [0,1]^n \to \mathbb{R} $ can be represented as:

\begin{equation}
f(x_{1},\dots,x_{n}) =
  \sum_{q=1}^{2n+1}
    \Phi_{q}\!\left(
      \sum_{p=1}^{n} \phi_{q,p}(x_{p})
    \right),
\end{equation}
where $\phi_{q,p} : [0,1] \to \mathbb{R}$, and $\Phi_{q} : \mathbb{R} \to \mathbb{R} $ are continuous univariate functions. There has been a long debate~\citep{girosi1989representation, schmidt2021kolmogorov} on the usefulness of this theorem for machine learning because of the general non-smoothness and wild behavior of the inner functions. Nevertheless, it inspired the construction of Kolmogorov-Arnold Networks~\citep{hecht1987kolmogorov, igelnik2003kolmogorov}, whose layers are defined as $y_q = \sum_p f_{qp}(x_p)$, where $f_{qp}$ are trainable functions. \citet{liu2024kan} introduced the modern version, which suggests stacking an arbitrarily large number of KAN layers and using an arbitrarily large number of neurons, similarly to MLPs. While~\citet{liu2024kan} illustrated strong performance of KANs, many test cases involve ground-truth functions with known, reasonably smooth KART or KART-like (matching KANs with more than one hidden layer and a larger number of neurons) closed-form representations. Subsequent works such as~\citet{yang2024kolmogorov}, \citet{kundu2024kanqas}, and \citet{kashefi2025pointnet} further reinforced the efficiency of KANs. On the contrary, \citet{yu2024kan} found that KANs can fall short compared to MLPs for some tasks.

The idea of lookup-based $\mathcal{O}(1)$ computations of KAN univariate functions is sometimes briefly mentioned but rarely implemented in practice~\citep{somvanshi2024survey, ji2024comprehensive}, likely because of challenges associated with an efficient GPU implementation. Surprisingly, most of the research goes in the somewhat opposite direction. B-splines, piecewise polynomial basis functions used in the original KAN paper, have compact support and thus are well suited for $\mathcal{O}(1)$ inference. Subsequent works often replace them with dense basis functions, such as Chebyshev polynomials~\citep{ss2024chebyshev} or Fourier harmonics~\citep{xu2024fourierkan}. The case of FastKAN~\citep{li2024kolmogorovarnold}, which replaces sparse B-splines with similar-looking dense Gaussian radial basis functions exclusively for the sake of optimization, is especially notable. 

A few works, such as \citet{moradzadeh2024ukan} and \citet{huang2025hardware}, implement the lookup idea. \citet{moradzadeh2024ukan}, however, predict B-spline coefficients using an MLP for the given grid points, which, thus, are not fully independent of each other. \citet{huang2025hardware} achieve remarkable efficiency on a small-scale problem from the original KAN paper by algorithm-hardware co-design using the TSMC 22 nm RRAM-ACIM chip. \citet{poluektov2025construction} and \citet{polar2021deep} employ piecewise linear parametrization suitable for $\mathcal{O}(1)$ inference but do not focus on inference efficiency.

In this work, we provide CUDA kernels for efficient inference and benchmark the introduced models against MLPs on general tasks where KART representations are not known in closed form and where there is no reason to expect them to be smoother than in other cases. 

\section{Lookup multivariate Kolmogorov-Arnold Networks}
\label{sec:lmkans}
At first glance, given that the inference cost of spline lookup tables does not depend on the number of parameters, very expressive univariate functions with tens of thousands of trainable parameters each are an ideal match for the Kolmogorov-Arnold Representation Theorem. However, KANs rarely use more than a few dozen parameters per function in practice. The difference between a univariate function parametrized by tens of thousands of parameters and just a few dozen is the capability of the former to parametrize a very high frequency band. On the one hand, this expressivity is necessary for closely approximating the 'wild behavior' of KART inner functions, but on the other, it raises concerns about training stability and generalization. On the contrary, multivariate functions can "accommodate" a significantly larger number of parameters without spilling expressive power into exceedingly high frequency bands. For instance, a four-dimensional function with just 10 grid points along each dimension has roughly the same number of trainable parameters as a univariate one with $\sim10^4$ grid intervals. 

\begin{figure}[H]
    \centering
    \includegraphics[width=0.5\linewidth]{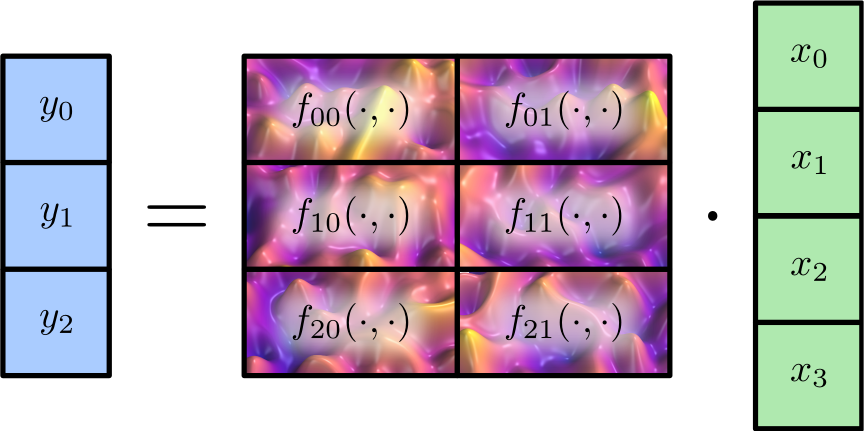}
    \caption{Schematic representation of a 2D lmKAN layer with 4 inputs and 3 outputs. This layer defines outputs as: $y_0 = f_{00}(x_0, x_1) + f_{01}(x_2, x_3)$, $y_1 = f_{10}(x_0, x_1) + f_{11}(x_2, x_3)$, and $y_2 = f_{20}(x_0, x_1) + f_{21}(x_2, x_3)$. The functions $f_{**}(\cdot, \cdot)$ are to be trained during fitting.}
    \label{fig:lmKAN}
\end{figure}

A layer of a multivariate version of Kolmogorov-Arnold Networks with dimension $d$ defines the output as:
\begin{equation}
    y_q = \sum_{p=0}^{N_{\text{in}}/d - 1} f_{qp}(x_{dp}, x_{dp + 1}..., x_{dp + d - 1})\text{,}
\end{equation}
where $f_{qp}$ are trainable d-dimensional functions and $N_{\text{in}}$ is the input dimensionality (assumed to be divisible by $d$). We implement CUDA kernels for the two-dimensional case. The motivation behind this choice is detailed in Sec.~\ref{sec:perspective}. An example of such a layer is depicted in Fig.~\ref{fig:lmKAN}. Similar to KANs, these layers do not need additional activations in between and can be stacked arbitrarily, substituting linear mapping-activation pairs in MLP-based backbones. 

In Sec.~\ref{sec:fastkan_comparison}, we empirically compare the two-dimensional version of lmKANs with one-dimensional FastKAN. The outcomes of these numerical experiments reinforce the intuitive considerations given here and suggest that multidimensional building blocks can indeed be more effective hosts for a large number of parameters in a practical setup. 

Additionally, it is worth noting that, if necessary, multivariate functions $f_{qp}$ can always fall back to sums of univariate ones, which would make the whole lmKAN fall back to standard KAN. Thus, the Kolmogorov-Arnold Representation Theorem is applicable also to our construction. 

\subsection{Function parametrization}
\label{sec:functions_parametrization}
\begin{figure}[h]
    \centering
    \includegraphics[width=0.9\linewidth]{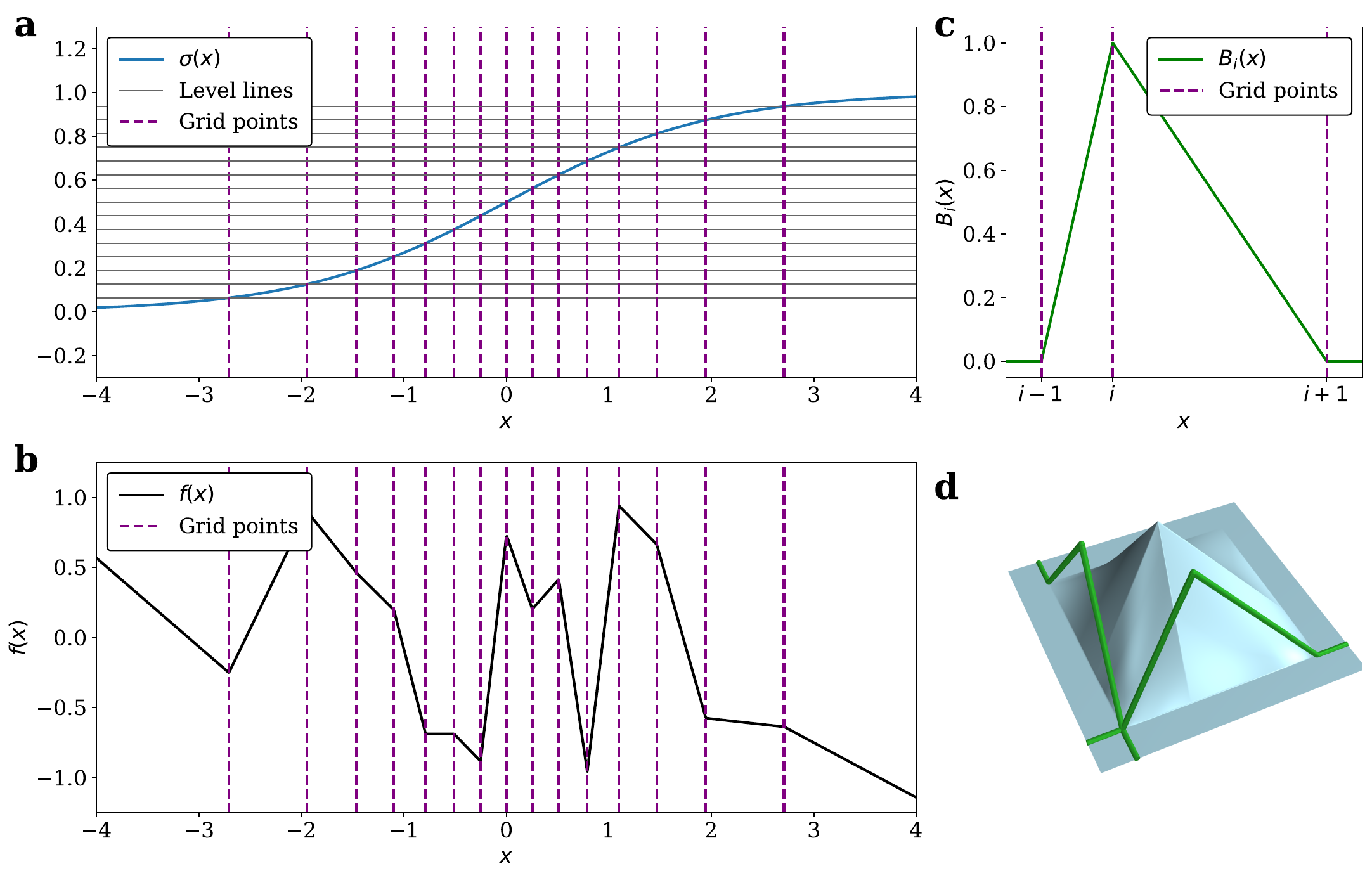}
    \caption{(a) Construction of the sigma grid; (b) an example of a piecewise linear function defined on such a grid (c) Second-order B-spline (d) Two-dimensional second-order B-spline}.
    \label{fig:cdf_grid}
\end{figure}

During training, activations of neurons can evolve arbitrarily, making the use of grids defined on bounded regions challenging. Therefore, we designed an unbounded grid which is still regular enough to allow $\mathcal{O}(1)$ computations. 
\paragraph{Sigma grid}
The one-dimensional sigma grid, which is illustrated in Fig.~\ref{fig:cdf_grid}a, is generated by any sigmoid-like function $\sigma(x)$. If the desired number of grid intervals is $G$, then the grid points are given as the intersection of $ G-1$ equispaced percentile levels with $\sigma(x)$. An example of a piecewise linear function defined on such a grid is given in Fig.~\ref{fig:cdf_grid}b. Such a construction spans the entire real axis. The grid has the finest resolution near the origin, and becomes progressively coarser as $|x|$ increases. For a given $x$, the index of the corresponding grid interval can be computed as $i = \lfloor \sigma(x) G \rfloor$, which makes such a grid suitable for $\mathcal{O}(1)$ computations. 

\paragraph{Static percentile grid} 
The choice of $\sigma(x)$ in the construction above is somewhat arbitrary. An additional consideration is that it would be beneficial to distribute inputs evenly, or approximately evenly, across the grid intervals. 
If the probability distribution of inputs is known, then one way to ensure perfect balance is to set $\sigma(x)$ to be the corresponding cumulative distribution function (CDF). However, it is challenging to implement such a \emph{dynamic percentile grid} in practice, as the distribution of inputs is unknown, can evolve during training, and the corresponding CDF can hardly be queried in $\mathcal{O}(1)$ time. 

Instead, we keep activations in a controlled range. A batch-normalization layer \emph{without affine parameters} placed before each lmKAN layer forces zero mean and unit variance. While it does not pin the entire distributions, it is safe to assume that they will be close enough to the standard normal for a reasonable ratio of neurons. Therefore, a viable solution would be to precede each lmKAN layer with a batch normalization with disabled affine transforms, and select $\sigma(x)$ to be the standard Gaussian CDF. In practice, we implement a fast approximation that requires computing only a single exponent function at inference, see Appendix~\ref{appendix:func_parametrization}. For multivariate functions, we apply the same one-dimensional grid independently to each coordinate.


\paragraph{B-splines}
We use B-splines of second order built on top of the described grids as basis functions to parametrize the lmKAN inner functions. A one-dimensional second-order B-spline centered around grid point $i$ is given in Fig.~\ref{fig:cdf_grid}c. It takes non-zero values on only two adjacent grid intervals around the center grid point. If there are $G$ intervals, then we use $G + 1$ basis functions: $G - 1$ such B-splines centered around each inner grid point, and two linear functions on the left-most and right-most infinite intervals. For any given point $x$, there are only two non-zero B-splines $B_i(x)$ - the ones with $i = \lfloor \sigma(x) G\rfloor$ and $i = \lfloor \sigma(x) G\rfloor + 1$ (In the corner case when $x$ coincides with one of the grid points, there is only one). Appendix~\ref{appendix:func_parametrization} provides the definition of B-splines on edge intervals we use in this work, along with other details.

A two-dimensional B-spline, illustrated in Fig.~\ref{fig:cdf_grid}d, is defined as $B_{i_1i_2}(x_1,x_2) = B_{i_1}(x_1)\,B_{i_2}(x_2)$.

All the functions defining a 2D lmKAN layer are parametrized as:
\begin{equation}
    f(x_1,x_2) = \sum_{i_1,i_2} p_{i_1i_2}\,B_{i_1i_2}(x_1,x_2)\text{,}
\label{eq:func_parametrization_2d}
\end{equation}
where $p_{i_1i_2}$ are trainable coefficients.

With such a construction, there are $(G + 1)^2$ independent parameters for each 2D function, parametrized functions are bilinear on each 2D grid interval, all the functions are continuous for arbitrary $p_{i_1i_2}$, and all but the edge coefficients $p_{i_1i_2}$ have a simple interpretation as the value of the function on the corresponding grid point. 

For any given point $(x_1, x_2)$, there are only four non-zero B-splines; thus, one needs to evaluate only four terms of Eq.~\ref{eq:func_parametrization_2d} to compute $f(x_1,x_2)$, which forms the basis of $\mathcal{O}(1)$ computations. The full algorithm to compute a standalone two-dimensional function is given in Appendix~\ref{appendix:2d_function_computation}.
 
\subsection{Computational cost}
\label{sec:computational_cost}

\begin{algorithm}[th]
  \caption{Forward pass of a 2D lmKAN layer.}
  \label{algorithm:lmKAN}
  \begin{algorithmic}[1]
    \Require input vector $\mathbf{x}\in\mathbb{R}^{N_{\text{in}}}$,
            parameter tensor $\mathbf{P}\in
            \mathbb{R}^{\bigl[G+1,\;G+1,\;\nicefrac{N_{\text{in}}}{2},\;N_{\text{out}}\bigr]}$. \newline
            $\mathbf{P}[\,i_1,\,i_2,\,\textit{input\_index},\,\textit{output\_index}\,]$ is the value of $f_{\textit{input\_index},\,\textit{output\_index}}$ at the \newline $i_1-$, $i_2-$th grid point.
    \Ensure output vector $\mathbf{y}\in\mathbb{R}^{N_{\text{out}}}$

    \State $\mathbf{y}\gets\mathbf{0}$ 
    \For{$\textit{input\_index}=0$ \textbf{to} $\nicefrac{N_{\text{in}}}{2}-1$}

        \State $i_1,i_2,
        B_{i_1i_2}(x_1,x_2), B_{i_1+1\,i_2}(x_1,x_2),
        B_{i_1\,i_2+1}(x_1,x_2), B_{i_1+1\,i_2+1}(x_1,x_2) \gets$
\Statex\hspace*{\dimexpr 60pt\relax}
       $\textbf{Preamble}\bigl(
         \mathbf{x}[2\!\cdot\!\textit{input\_index}],
         \mathbf{x}[2\!\cdot\!\textit{input\_index}+1]\bigr)$

        \For{$\textit{output\_index}=0$ \textbf{to} $N_{\text{out}}-1$}
            \State {$\mathbf{y}[\textit{output\_index}] \mathrel{+}= B_{i_1\,i_2}(x_1, x_2)\;
                   \mathbf{P}[\,i_1,\,i_2,\,\textit{input\_index},\,\textit{output\_index}\,]$}
            \State {$\mathbf{y}[\textit{output\_index}] \mathrel{+}= B_{i_1+1\,i_2}(x_1, x_2)\;
                   \mathbf{P}[\,i_1\!+\!1,\,i_2,\,\textit{input\_index},\,\textit{output\_index}\,]$}
            \State {$\mathbf{y}[\textit{output\_index}] \mathrel{+}= B_{i_1\,i_2+1}(x_1, x_2)\;
                   \mathbf{P}[\,i_1,\,i_2\!+\!1,\,\textit{input\_index},\,\textit{output\_index}\,]$}
            \State {$\mathbf{y}[\textit{output\_index}] \mathrel{+}= B_{i_1+1\,i_2+1}(x_1, x_2)\;
                   \mathbf{P}[\,i_1\!+\!1,\,i_2\!+\!1,\,\textit{input\_index},\,\textit{output\_index}\,]$}
        \EndFor
    \EndFor

    \Statex
    \Function{\textbf{Preamble}}{$x_1,x_2$}
      \Comment{Handling of edge-interval cases is omitted for brevity.  See Appendix~\ref{appendix:func_parametrization} for more details.}
      \Require $x_1$, $x_2$; precomputed grid points $\bm{\mathcal{G}}$; \newline
              precomputed inverse areas $\mathbf{A}_{\text{inv}}[i_1, i_2] = \bigl[(\bm{\mathcal{G}}[i_1 + 1] - \bm{\mathcal{G}}[i_1]) (\bm{\mathcal{G}}[i_2 + 1] - \bm{\mathcal{G}}[i_2])\bigr]^{-1}$
      \Ensure indices $i_1,i_2$ and \newline B-splines $B_{i_1\,i_2}(x_1, x_2),B_{i_1+1\,i_2}(x_1, x_2),B_{i_1\,i_2+1}(x_1, x_2),  B_{i_1+1\,i_2+1}(x_1, x_2)$
      \State {$i_1 \gets \bigl\lfloor \sigma(x_1)\,G \bigr\rfloor$}
      \State {$i_2 \gets \bigl\lfloor \sigma(x_2)\,G \bigr\rfloor$}
      \State {$B_{i_1\,i_2}(x_1, x_2) \gets (\bm{\mathcal{G}}[i_1+1] - x_1)\,
                         (\bm{\mathcal{G}}[i_2+1] - x_2)\,
                         \mathbf{A}_{\text{inv}}[i_1,i_2]$}
       \State {$B_{i_1+1\,i_2}(x_1, x_2) \gets (x_1 - \bm{\mathcal{G}}[i_1])\,
                         (\bm{\mathcal{G}}[i_2+1] - x_2)\,
                         \mathbf{A}_{\text{inv}}[i_1,i_2]$}
        \State {$B_{i_1\,i_2+1}(x_1, x_2)\gets (\bm{\mathcal{G}}[i_1+1] - x_1)\,
                         (x_2 - \bm{\mathcal{G}}[i_2])\,
                         \mathbf{A}_{\text{inv}}[i_1,i_2]$}
      \State {$B_{i_1+1\,i_2+1}(x_1, x_2) \gets (x_1 - \bm{\mathcal{G}}[i_1])\,
                         (x_2 - \bm{\mathcal{G}}[i_2])\,
                         \mathbf{A}_{\text{inv}}[i_1,i_2]$}

      \State \Return $i_1,i_2,B_{i_1\,i_2}(x_1, x_2),B_{i_1+1\,i_2}(x_1, x_2),B_{i_1\,i_2+1}(x_1, x_2),B_{i_1+1\,i_2+1}(x_1, x_2) $
    \EndFunction
  \end{algorithmic}
\end{algorithm}

Overall, the functional form of lmKANs involves computations of many low-dimensional functions with exactly the same arguments (those in the same column, see Fig.~\ref{fig:lmKAN}). When doing so, it is possible to reuse many of the intermediate values, such as indices of the grid intervals and B-splines. As Algorithm~\ref{algorithm:lmKAN} elaborates, these intermediate values can be computed once for each pair of inputs, and then utilized to compute the value of a given function with just four multiply-add operations for the 2D case.

As the algorithm shows, the preamble part contributes only to an asymptotically insignificant $\mathcal{O}(N)$ term, where $N$ is the input ($N_{\text{in}}$) or output ($N_{\text{out}}$) dimension. Given that the total number of 2D functions in an lmKAN layer is $[N_{\text{in}}/2] N_{\text{out}}$, the total number of required multiply-add operations for the dominant, $\mathcal{O}(N^2)$, part is $4 [N_{\text{in}}/2] N_{\text{out}} = 2 N_{\text{in}}N_{\text{out}}$, just $\mathbf{2\times}$ that of a linear layer of the same shape. Following the common practice~\cite{he2016deep}, we estimate FLOPs as the number of fused multiply-adds of the main asymptotic term for both MLPs and lmKANs. 

It is worth noting that the $\mathcal{O}(N)$ preamble term is not an additional cost relative to MLPs; it replaces the per-unit bias additions and activation evaluations that lmKANs do not require --- operations that can be quite expensive when the activations are transcendental functions such as \texttt{tanh}. 
⁡
\subsection{Perspective on dimensions and spline orders}
\label{sec:perspective}
All the constructions introduced so far straightforwardly generalize to a higher-dimensional case. Evaluation of a single d-dimensional function parametrized by d-dimensional second-order B-splines would take $2^d$ multiply-adds, which stem from the summation of B-spline contributions residing on all the corners of the corresponding d-dimensional hypercube. Given that the number of such d-dimensional functions would be $d$ times smaller compared to the number of weights of a linear layer of the same shape, the inference FLOPs of the d-dimensional lmKAN layer would be $2^d /d$ of that of the linear layer with the same shape. 

These slowdown factors are exactly identical, 2×, for one- and two-dimensional lmKANs, and start to grow for higher dimensions. Thus, we chose the two-dimensional version for the implementation of CUDA kernels, as this extension of the standard univariate KAN comes essentially for free. 

If B-splines of order $k$ are employed for parametrization instead of the second-order ones described so far, the inference cost becomes $k^d / d$ of that of a linear layer. For more details, see the B-spline definition at \citet{de1978practical}, and how they are used in KANs~\cite {liu2024kan}. Increasing the B-spline order brings a few benefits, but they likely do not justify the associated increase in the computational cost. 

The classical theorem about B-splines~\cite{de1968uniform} indicates that while the order of B-splines affects the convergence rate, any spline order is sufficient to approximate any function arbitrarily closely by increasing the resolution of the grid\footnote{This is applicable, though, to functions on a bounded domain.}. Given that the computational cost of spline look-up tables does not depend on the number of grid points, the grid resolution can be arbitrarily increased without any computational overhead at inference. 

On top of that, increasing the B-spline order introduces additional smoothness of the model. Functions parametrized by B-splines of order $k$ belong to $C^{k-2}$, but in general not to $C^{k-1}$. The smoothness of second-order B-splines employed in this work matches that of ReLU~\cite{glorot2011deep}, one of the most popular and successful activation functions, which is also continuous, but not continuously differentiable. Thus, it is questionable if the additional smoothness available through higher orders $k$ is necessary and would justify the associated computational overhead. 

\subsection{CUDA kernels}
In this work, we implement CUDA kernels for efficient inference of 2D lmKANs on modern GPUs. Our kernels use the classic shared-memory tiling used in GEMM~\cite{volkov2008benchmarking}. In the following, all benchmarks run in full-precision \texttt{float32}.

With a $16\times16$ tile, our implementation is $\sim8\times$ slower than a dense linear layer with the same shape on an H100-SXM, irrespective of the grid resolution. The slowdown exceeds the $\sim2\times$ FLOPs-based estimate from Sec.~\ref{sec:computational_cost}, primarily because of the less coherent memory-access pattern of algorithm~\ref {algorithm:lmKAN}. Additionally, dense matrix multiplication has been the cornerstone of many computational pipelines and, thus, has enjoyed decades of thorough optimization.

Finite shared-memory capacity limits the number of grid intervals to
$G\!\le20$ on H100.  At this limit, an lmKAN layer holds
$(20+1)^2/2\approx220\times$ more parameters than the linear baseline with the same shape, so \textbf{its inference time per trainable parameter is }$\mathbf{\bigl[(20+1)^2/2\bigr]/8\approx27.5\times}$\textbf{
better}.

Reducing the tile to $8\times8$ raises the slowdown to $\sim9.5\times$ but lets us increase $G$ to 40, yielding $\sim\mathbf{88.5\times}$ \textbf{better per-parameter efficiency.}

The performance of our kernels is even better when the feature dimension is small. The numbers reported above were obtained in the limit of large batch sizes and feature dimensions. Keeping the batch size large but setting, for instance,  $N_{\mathrm{in}} = N_{\mathrm{out}} = 32$ lowers the slowdown of the $16\times16$ kernel to only $\sim2.5\times$ relative to a linear layer of the same shape. Further details are reported in Appendix~\ref{appendix:cuda_kernels}. 

\subsection{Hessian regularization}
\label{sec:hessian_regularization}
\begin{figure}[H]
    \centering
    \includegraphics[width=1.0\linewidth]{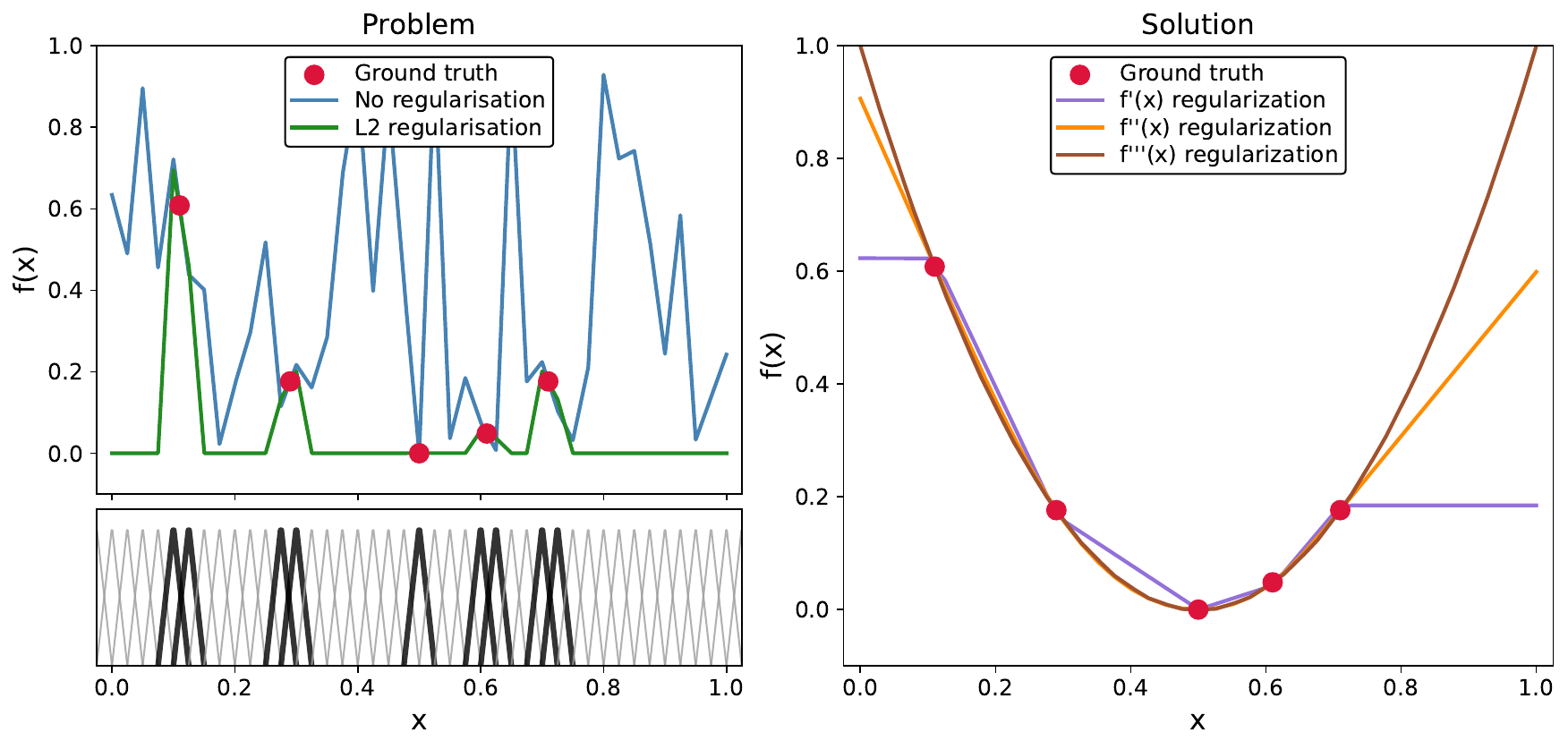}
    \caption{Generalization pitfall.}
    \label{fig:derivatives_regularization}
\end{figure}

Direct fitting of splined functions with a fine grid imposes additional challenges related to generalization. The problem is illustrated in Fig.~\ref{fig:derivatives_regularization}a for the simple case of fitting a standalone one-dimensional function parametrized by second-order B-splines on a uniform grid on [0, 1] with $G=40$ intervals, $f(x) = \sum_i p_i B_i(x)$. 

Ground truth is an exact parabola, and the training set consists of 5 points, which are illustrated on the plot. Since the number of grid points is large, the model has enough flexibility to reproduce the training set exactly, but generalization is quite off. With such a fine grid, only a few B-splines, marked as bold on the bottom panel of Fig.~\ref{fig:derivatives_regularization}a, take non-zero values for the training points. Thus, only the coefficients $p_i$ associated with these active B-splines receive non-zero gradient during training, while all the others have no incentive to evolve from the random values assigned at initialization. Standard L2 regularization is not much better, as it simply pushes all non-active coefficients $p_i$ to zero, which also results in a non-meaningful approximation after training. 

When dealing with a similar problem, \citet{xie2023ultra} employed off-diagonal regularization based on a finite-difference scheme for the second derivative. One can put regularization terms as $\lambda \sum_i(p_i - p_{i + 1})^2$ for first derivative, $\lambda \sum_i(p_i - 2 p_{i + 1} + p_{i + 2})^2$ for second, and so on. Such regularization schemes result in meaningful approximations after training, as illustrated in Fig.~\ref{fig:derivatives_regularization}b.

We implemented CUDA kernels for 2D lmKANs, which express general high-dimensional mapping in terms of building blocks of two-dimensional functions. For 2D functions, we use off-diagonal regularization based on the squared Frobenius norm of the Hessian, a rotationally invariant measure of curvature in any direction, which is not to be confused with the Laplacian. Furthermore, our finite-differences schemes take into account that the grids, introduced in Sec.~\ref{sec:functions_parametrization}, are not uniform. More details are given in Appendix~\ref{appendix:off_diagonal_regularization}. 

The Hessian of a function is zero if and only if the function is linear. Therefore, the use of a very strong Hessian-based regularization leads to linearization of the trained functions, enforcing them to converge to $f(x_1, x_2) = a x_1 + bx_2 + c$. This makes the whole lmKAN equivalent to an MLP of the same shape, modulo training dynamics. In other words, the Hessian regularization coefficient $\lambda$ can be used to smoothly adjust the lmKAN behavior between fully unconstrained lmKAN and MLP extremes. This observation—that lmKANs with heavy Hessian regularization match non-regularized MLPs—suggests that one should use a combination of the proposed regularization scheme with standard ones, such as L2 or dropout~\cite{srivastava2014dropout}, for the best results. 

\paragraph{Preconditioning and fitting scheme} Similarly to the original KANs~\cite{liu2024kan}, we append a preconditioning term to splined functions, which in our case is linear. To further increase training stability, we employ a multi-staged fitting procedure, where the strength of the described Hessian regularization is initially set to a high value and then gradually decays. More details are available in Appendix~\ref{appendix:preconditioning}. 

\section{Experiments}
\label{sec:experiments}
We have demonstrated so far that lmKANs can have significantly better inference cost per trainable parameter compared to linear layers in terms of both FLOPs and wall-clock time on modern GPUs. The question is, however, whether this nominal efficiency translates to real-life performance. Do lmKANs indeed represent a better trade-off between performance and inference cost? 

In this section, we empirically compare the efficiency of lmKANs and MLPs across the following settings: (i) approximating general high-dimensional functions, (ii) on a tabular-like dataset of randomly displaced methane configurations, and (iii) within CNN frameworks evaluated on CIFAR-10 and ImageNet. Across all experiments, we use identical macro-architectural backbones for lmKANs and MLPs. Overall, to obtain a comprehensive picture of the performance, we prioritized the diversity of the setups over a very large scale or the architectural complexity of a particular backbone. We found that lmKANs are consistently inference FLOPs --- accuracy Pareto-optimal, with the largest gains on the methane dataset. Finally, we compare lmKANs with FastKANs. 

\subsection{General function approximation}
\label{sec:general_function_approximation}
Our first experiment is set to measure crude flexibility of lmKANs in approximating general high-dimensional functions, which we model by large teacher MLPs with fixed random weights. We define a ground-truth $\mathbb{R}^{32} \rightarrow \mathbb{R}^{1}$ function as an MLP with $32$ input neurons, $10$ hidden layers, each with $1024$ neurons, and hyperbolic tangent activations. When weights are initialized using the default \texttt{PyTorch} initialization, the magnitude of activations progressively decreases from layer to layer. Thus, to avoid this, we multiplied all the weights by $3.0$ after random initialization. 

\begin{figure}[th]
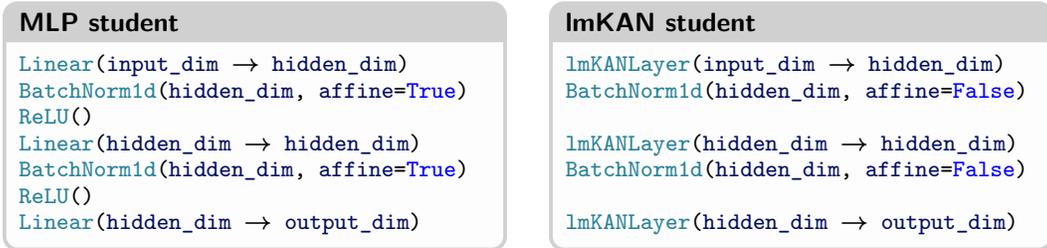

  \centering
  \begin{minipage}[t]{0.48\linewidth}
    \begin{archbox}{MLP student}
Linear(input_dim -> hidden_dim)
BatchNorm1d(hidden_dim, affine=True)
ReLU()
Linear(hidden_dim -> hidden_dim)
BatchNorm1d(hidden_dim, affine=True)
ReLU()
Linear(hidden_dim -> output_dim)
    \end{archbox}
  \end{minipage}\hfill
  \begin{minipage}[t]{0.48\linewidth}
    \begin{archbox}{lmKAN student}
lmKANLayer(input_dim -> hidden_dim)
BatchNorm1d(hidden_dim, affine=False)

lmKANLayer(hidden_dim -> hidden_dim)
BatchNorm1d(hidden_dim, affine=False)

lmKANLayer(hidden_dim -> output_dim)
    \end{archbox}
  \end{minipage}

  \caption{Pseudocode for MLP and lmKAN students.}
  \label{fig:pseudocode}
\end{figure}

We fit both MLP and lmKAN students to approximate this ground-truth function and compare their performance. We use the same fully connected backbone for both types of models with two hidden layers and varying hidden dimensions. Both MLPs and lmKANs use batch normalizations. We set \texttt{affine=True} for MLPs as it is the standard choice, and \texttt{affine=False} for lmKANs in accordance with static percentile grids introduced in Sec.~\ref{sec:functions_parametrization}. MLPs use ReLU activations, while lmKANs do not require any additional activation functions. We use $G = 12$ for all the lmKAN models, as this was the optimal value found in the ablation study described below. Pseudocode for both models is available in Fig.~\ref{fig:pseudocode}. Both students have two hidden layers, which is one more than both Cybenko~\cite{cybenko1989approximation} (the one for MLPs) and Kolmogorov-Arnold universal approximation theorems require. This setup, however, is more realistic, as MLPs with exactly one hidden layer are rarely used in practice. 

We fit all the models with the Adam optimizer~\cite{kingma2014adam}. For each step of stochastic gradient descent, we generate random inputs from the normal distribution, and compute the corresponding targets by evaluating the ground-truth teacher MLP. In such an infinite data regime, the final Mean-Squared Error (MSE) depends on how flexible the models are, and monotonically decreases with the hidden dimension for both MLPs and lmKANs. 

\begin{figure}[H]
    \centering
    \includegraphics[width=\linewidth]{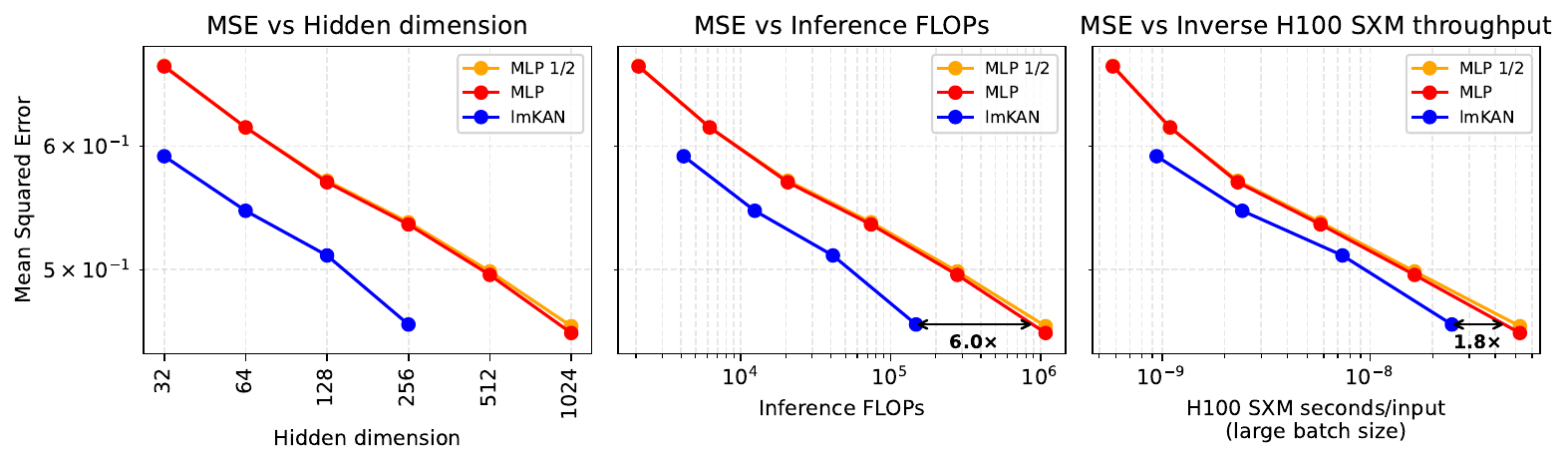}
    \caption{lmKAN vs MLP for general function approximation. The "MLP 1/2" line corresponds to the outcome of the fitting procedure with only half of the training steps compared to the "MLP" one.}
    \label{fig:general_function_approximation}
\end{figure}

The increase of the hidden dimension inevitably entails a higher computational cost at inference, and the question is which family of models represents a better trade-off between accuracy and computational efficiency. Our findings are summarized in Fig.~\ref{fig:general_function_approximation}. The left column illustrates the final MSE of converged models depending on the hidden dimension, the middle column represents the Pareto front between the final MSE and FLOPs required at inference, and the last column contains the Pareto front between the final MSE and inference H100 wall-clock time. 

In order to justify the claims that lmKANs are indeed more efficient, we converge baseline MLP-based models very tightly here, and in all similar experiments in this manuscript. For the MLP baseline, we have two lines - one with a full training budget and one with only half of it. The fact that these lines nearly coincide with each other demonstrates very tight convergence. 

Fig.~\ref{fig:general_function_approximation} clearly indicates that lmKANs are significantly more FLOPs efficient at the same accuracy level, up to \textbf{6×}, for the largest dimensions. Furthermore, lmKANs also appeared to be H100 wall-clock time optimal for all the scales, with the speed-up factor of about 1.8× for the largest hidden dimension. 

As an additional experiment, we fitted the same MLP and lmKAN students to approximate an $\mathbb{R}^{32} \rightarrow \mathbb{R}^{32}$ function represented by a similar ground-truth MLP with random weights. For this setup, lmKANs also appeared to be Pareto optimal, with FLOPs reduction up to 4.2×. More details are available in Appendix~\ref{appendix:general_function_approximation}. 

\paragraph{Ablation} 

\begin{figure}[h] 
        \centering
        \includegraphics[width=0.7\linewidth]{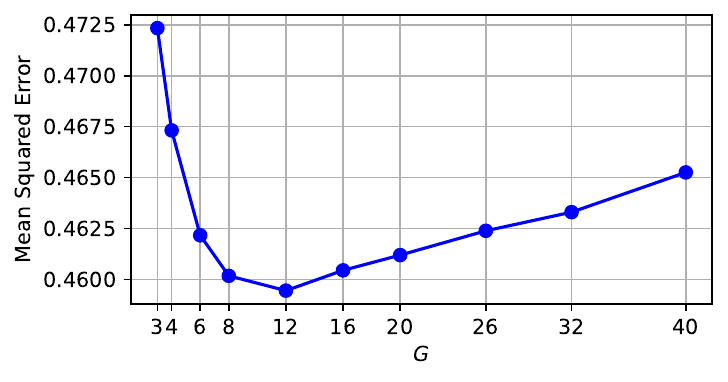}
        \caption{Final MSE vs $G$ for the hidden\_dim=256 lmKAN model.}
        \label{fig:general_function_n_chunks_plot}
\end{figure}

We investigate the effect of the chosen number of grid intervals $G$ on the resulting lmKAN accuracy when approximating the $\mathbb{R}^{32} \rightarrow \mathbb{R}^{1}$ function with hidden\_dim = $256$. The result is given in Fig.~\ref{fig:general_function_n_chunks_plot}.

Contrary to our initial expectations, the final MSE does not monotonically decrease with the grid resolution, having a distinct minimum at $G=12$. This is happening despite the fact that our setup effectively corresponds to an infinite data regime, effectively ruling out overfitting as an explanation. We hypothesize that the reason is that Kolmogorov-Arnold Networks are hard to converge for excessively fine grid resolution. Therefore, the optimal value of $G$ could depend on the training protocol. Appendix~\ref{appendix:general_function_approximation} reinforces this supposition by providing the MSE vs. epoch number plot with several lines corresponding to different $G$ values. Additionally, note that FastKANs were found to display the same effect to an even greater degree, as detailed in Sec.~\ref{sec:fastkan_comparison}.

\subsection{Randomly displaced methane configurations}
\label{sec:methane}

 \begin{wrapfigure}{r}{0.3\linewidth} 
  \vspace{-1.0\baselineskip}                
  \centering
  \includegraphics[width=\linewidth]{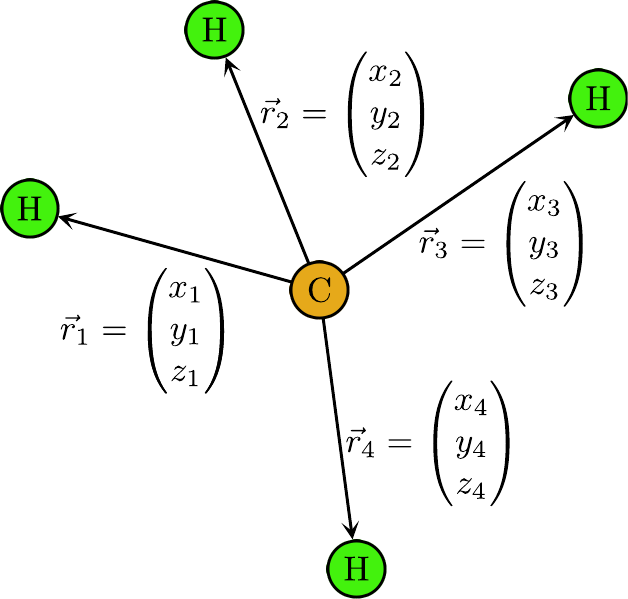}
  \caption{A methane configuration}
  \label{fig:methane}
\end{wrapfigure}

Our next step was to benchmark the performance of lmKANs on real data. Tabular datasets are the natural choice for feedforward fully connected neural networks. Popular tabular datasets, such as the Titanic~\cite{KaggleTitanic} or housing prices~\cite{harrison1978hedonic}, however, are not particularly convenient for this purpose. First, they are typically stochastic in nature - for instance, while it is possible to improve a guess on the survival based on the data available for the Titanic dataset, it is impossible to say for sure. Thus, even an arbitrarily large model fitted on arbitrarily many data points would have a non-zero limitation on the accuracy. In other words, the performance of a model translates into an error metric not so directly, making the comparisons between different models less illustrative. Second, these datasets are typically relatively small, making it challenging to sweep across a wide range of model scales to obtain a comprehensive picture of performance. 

Therefore, we chose the tabular-like dataset of randomly displaced methane configurations~\cite{pozdnyakov2020incompleteness} for the comparison. It consists of multiple off-equilibrium methane configurations, as illustrated in Fig.~\ref{fig:methane}. The target is given by the corresponding quantum-mechanical energy~\cite{turney2012psi4, kohn1965self}. Hydrogen atoms are placed around the carbon atom not as an ideal tetrahedron of an equilibrium methane molecule, but rather randomly, varying from instance to instance. Thus, the corresponding quantum-mechanical energies of such configurations are also different from each other. Machine learning models fitted on such datasets belong to the class of so-called machine learning interatomic potentials~\cite{behler2007generalized, bartok2010gaussian}. This dataset is sufficiently large for the comparisons, containing more than seven million configurations. Additionally, this dataset is deterministic - the geometry of the corresponding methane configuration completely determines the target (Formally, there can be a stochastic term due to the lack of complete convergence of ab initio computations for the quantum-mechanical energy, but it is negligible in practice). 

The target, the potential energy of the system, is invariant with respect to rotations and permutations of identical atoms\footnote{Formally, there is an additional symmetry, inversion, but the corresponding group contains only two elements, thus it does not make much sense to treat it separately. We treat it as part of the rotation group, and in the following, by rotation we mean proper or improper rotation. }. Therefore, there are several viable representations of the methane molecules depending on how these symmetries are addressed:

{\ttfamily\bfseries Cartesian Components:} The simplest representation is a collection of all the Cartesian components of all displacement vectors from the carbon atom to all the hydrogen atoms. Since each methane molecule contains $4$ hydrogen atoms, the total number of displacement vectors is $4$, and the total number of components is $12$. When using this representation, we simply concatenate all these components together and feed them to a fully connected MLP or lmKAN whose input dimension is $12$. This representation is not invariant with respect to both rotations and permutations; thus, we use the corresponding augmentations during training. We randomly permute hydrogen atoms and rotate each molecule whenever we sample a minibatch from the training subset for each step of stochastic gradient descent.

{\ttfamily\bfseries Distances:} Another possible representation is a collection of all the interatomic distances between all the atoms. Since the total number of atoms is $5$, the number of all the interatomic distances is $5 * 4 / 2 = 10$. Therefore, the input dimension of fully connected networks applied to this representation is $10$. This representation is invariant with respect to rotations but not with respect to permutations. During training, we use only permutational augmentations. 

{\ttfamily\bfseries Cartesian Components Polynomials:} We compute power sum symmetric polynomials on top of the Cartesian components of the displacement vectors: $P_{\alpha_x, \alpha_y, \alpha_z} = \sum_{i = 1}^{i=4} x_i^{\alpha_x} y_i^{\alpha_y} z_i^{\alpha_z}$ for non-negative integer $\alpha_x + \alpha_y + \alpha_z \leq 4$. The total number of such symmetric polynomials is $34$ (excluding trivial $P_{0, 0, 0}$). This representation is invariant with respect to permutations but not with respect to rotations. Thus, during training we use only rotational augmentations. 

{\ttfamily\bfseries Distances Polynomials:} The final representation is a collection of non-trivial symmetric polynomials on top of the interatomic distances, constructed similarly as in~\citet{allen2021atomic}. The total number of such polynomials is $31$, and their exact formulas are given in Appendix~\ref{appendix:methane}. This representation is invariant with respect to both rotations and permutations. Thus, we do not use any augmentations during training for this representation. 

\begin{table}[t]
  \caption{Summary of methane representations}
  \label{tab:methane_representations}
  \centering
  \renewcommand{\arraystretch}{1.3}
  \begin{tabular}{cccc}
    \toprule
    Label & \shortstack{Rotational\\symmetry} & \shortstack{Permutational\\symmetry} & \#Features \\
    \midrule
    {\ttfamily\bfseries Cartesian Components} & Augmentations & Augmentations & 12 \\
    {\ttfamily\bfseries Distances} & \shortstack{Features} & Augmentations & 10 \\
    \shortstack{{\ttfamily\bfseries Cartesian Components Polynomials}} & Augmentations & \shortstack{Features} & 34 \\
    \shortstack{{\ttfamily\bfseries Distances Polynomials}} & \shortstack{Features} & \shortstack{Features} & 31 \\
    \bottomrule
  \end{tabular}
\end{table}

The described representations are summarized in Table~\ref{tab:methane_representations}. We systematically evaluate all four possible combinations of how the rotational and permutational symmetries can be incorporated into the fitting pipeline. Within the {\ttfamily\bfseries Distances Polynomials} representation, the methane dataset is tabular in the classical sense --- it is a table with about 7.7~million rows and 31 columns. For other representations, the dataset is tabular-like given the available augmentation strategies. We randomly split the data into \texttt{7000000}, \texttt{300000}, and \texttt{432488} train, validation, and test molecules, respectively. 

\begin{figure}[h]
    \centering
    \includegraphics[width=\linewidth]{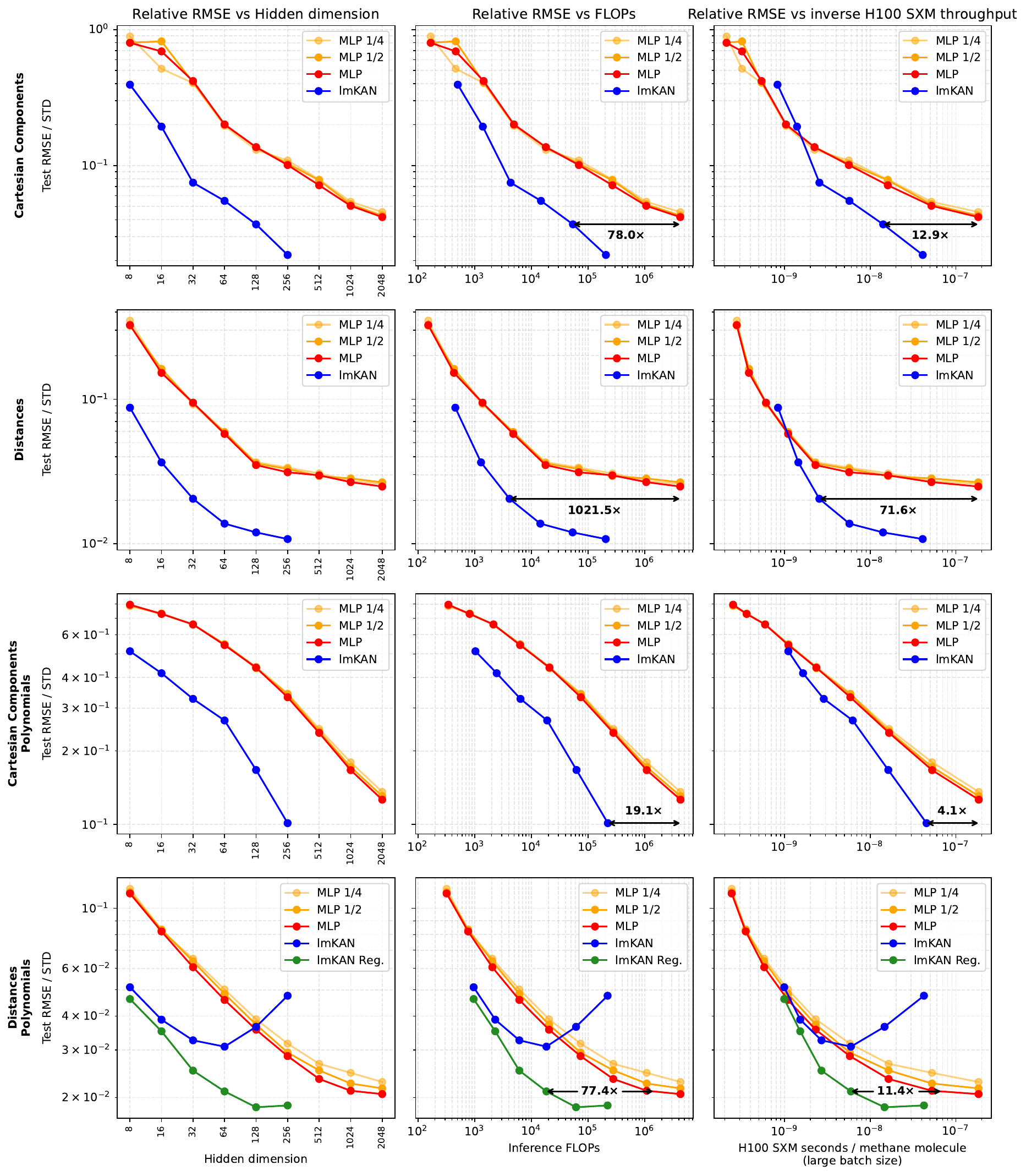}
    \caption{lmKAN vs MLP on the dataset of randomly displaced methane configurations. "lmKAN Reg." curve corresponds to lmKAN fitted with Hessian regularization introduced in Sec.~\ref{sec:hessian_regularization}. On the vertical axis, we plot the relative Root Mean Squared Error, which is given as test RMSE normalized by standard deviation of the target in the dataset. The "MLP 1/2" and "MLP 1/4" curves correspond to outcomes of fitting procedures with half and a quarter of the training budget, respectively. }
    \label{fig:methane_results}
\end{figure}

For each representation, we fit the same families of MLP and lmKAN models as in the previous section. The result is given in Fig.~\ref{fig:methane_results}. For this dataset, we use $G=28$, the optimal value we found in ablation studies. Similarly to the previous experiment, we demonstrate tight convergence of the baseline MLP models by providing three lines corresponding to full, half, and quarter of the training budget, respectively. Overall, when compared to domain-specific architectures, typically given by GNNs~\cite{zhang2021physically} and/or transformers~\cite{pozdnyakov2023smooth}, the introduced feedforward fully connected models occupy a non-overlapping part of the Pareto frontier --- they are less accurate, but also orders of magnitude faster.

The figure illustrates that lmKANs consistently outperform MLPs across all modalities. Furthermore, the performance improvement is much larger compared to our previous experiment. At the same accuracy level, lmKANs require dozens of times (or even up to one thousand for the {\ttfamily\bfseries Distances} modality) fewer inference FLOPs, which results in \textbf{more than a 10×} improvement of the inference H100 wall-clock time. 

On top of that, Fig.~\ref{fig:methane_results} provides early indications that lmKANs sometimes can be more accurate in the limit of large scale, that is, to have better generalizability. The second row of the figure, corresponding to the {\ttfamily\bfseries Distances}  modality, illustrates that the rate of improvement of MLP models becomes very slow, and it is questionable if this family of models would ever surpass the accuracy achieved by lmKAN models at any scale. 

On the other hand, depending on the nature of the data, raw lmKANs, without the Hessian regularization we proposed in Sec.~\ref{sec:hessian_regularization}, can be more prone to overfitting. This happens for the {\ttfamily\bfseries Distances Polynomials} modality as the last row of Fig.~\ref{fig:methane_results} illustrates. This modality incorporates all the symmetries into the representation and does not involve any sort of augmentations. Therefore, it is likely that the generalization problem we outlined in Sec.~\ref{sec:hessian_regularization} takes place for this fitting setup. As the green line of the fourth row of Fig.~\ref{fig:methane_results} illustrates, the Hessian regularization is sufficient to overcome the overfitting. Properly regularized lmKANs were found to outperform the MLPs and be Pareto optimal from the point of view of both inference FLOPs and inference H100 wall-clock time.

\paragraph{Ablations} Appendix~\ref{appendix:methane} contains ablation studies focusing on the grid resolution $G$ and the strength of Hessian regularization. 



\subsection{lmKAN-based Convolutional Neural Networks}
\label{sec:cifar_10}

In the introduction, we briefly outlined that high-dimensional linear mappings are the primary building blocks in most architectures, not only in feedforward fully connected neural networks. Convolutional Neural Networks are no exception. 

A standard computer-vision two-dimensional convolution with kernel size $k \times k$ is parametrized by linear mapping $\mathbb{R}^{k^2 C_{in}} \rightarrow \mathbb{R}^{C_{out}}$, where $C_{in}$ and $C_{out}$ are numbers of input and output channels, respectively. Since Kolmogorov-Arnold layers can be used as a general substitute for high-dimensional linear mappings, one can construct a KAN-based convolutional neural network well suited for image processing, as was done, e.g, in~\citet{bodner2024convolutional}. In this section, we compare the performance of lmKAN- and MLP-based CNNs on the CIFAR-10~\cite{krizhevsky2009learning} and ImageNet~\cite{deng2009imagenet} datasets.

\begin{figure}[th]
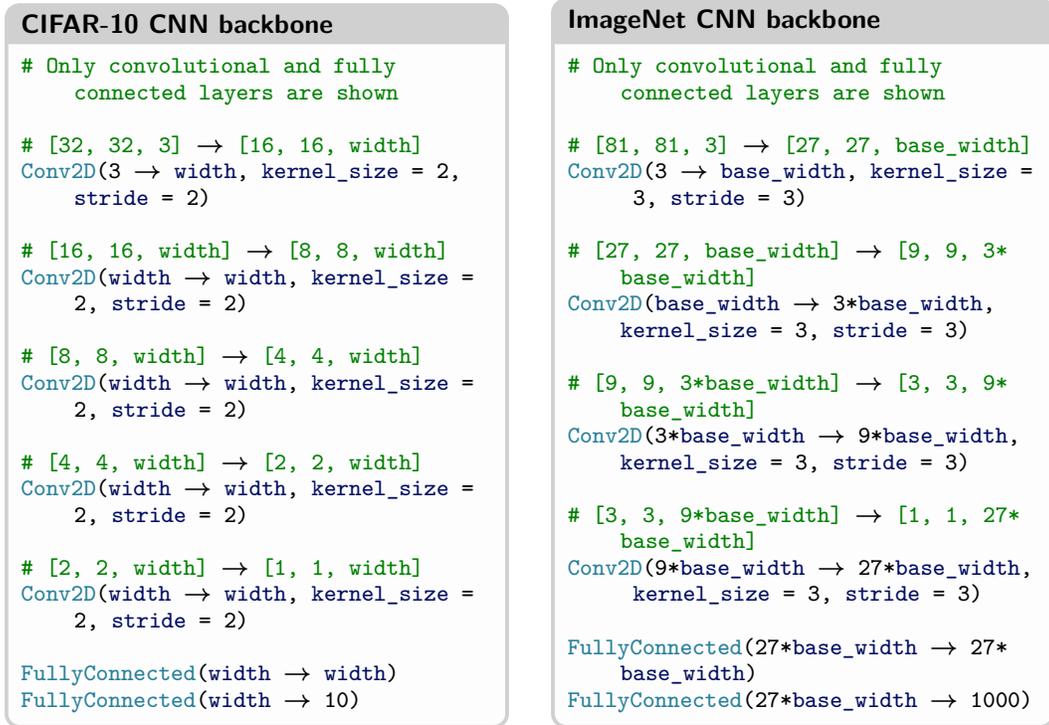

  \centering
  \begin{minipage}[t]{0.48\linewidth}
    \begin{archboxcnnX}{CIFAR-10 CNN backbone}
# Only convolutional and fully connected layers are shown

# [32, 32, 3] -> [16, 16, width]
Conv2D(3 -> width, kernel_size = 2, stride = 2)

# [16, 16, width] -> [8, 8, width]
Conv2D(width -> width, kernel_size = 2, stride = 2)

# [8, 8, width] -> [4, 4, width]
Conv2D(width -> width, kernel_size = 2, stride = 2)

# [4, 4, width] -> [2, 2, width]
Conv2D(width -> width, kernel_size = 2, stride = 2)

# [2, 2, width] -> [1, 1, width]
Conv2D(width -> width, kernel_size = 2, stride = 2)

FullyConnected(width -> width)
FullyConnected(width -> 10)
    \end{archboxcnnX}
  \end{minipage}\hfill
  \begin{minipage}[t]{0.48\linewidth}
    \begin{archboxcnnX}{ImageNet CNN backbone}
# Only convolutional and fully connected layers are shown

# [81, 81, 3] -> [27, 27, base_width]
Conv2D(3 -> base_width, kernel_size = 3, stride = 3)

# [27, 27, base_width] -> [9, 9, 3*base_width]
Conv2D(base_width -> 3*base_width, kernel_size = 3, stride = 3)

# [9, 9, 3*base_width] -> [3, 3, 9*base_width]
Conv2D(3*base_width -> 9*base_width, kernel_size = 3, stride = 3)

# [3, 3, 9*base_width] -> [1, 1, 27*base_width]
Conv2D(9*base_width -> 27*base_width, kernel_size = 3, stride = 3)

FullyConnected(27*base_width -> 27*base_width)
FullyConnected(27*base_width -> 1000)
    \end{archboxcnnX}
  \end{minipage}

  \caption{CIFAR-10 and ImageNet CNN backbones. MLP-based CNNs additionally have ReLU activations and batch normalizations with enabled affine transforms. lmKAN-based CNNs do not require additional activations and use batch normalizations without affine transforms as suggested by our static percentile grids described in Sec.~\ref{sec:functions_parametrization}.}
  \label{fig:cnn_backbones}
\end{figure}

\subsubsection{CIFAR-10} Our backbone architecture consists of five $2 \times 2$ convolutions, each with stride $2$, and two fully connected layers at the end. Since the resolution of CIFAR-10 images is $32\times32$, where $32=2^5$, five $2 \times 2$ convolutions with stride $2$ transform the spatial dimensions of an image exactly to $1\times1$. All the layers use the same width (= number of filters in case of convolutions, and hidden dimension in case of fully connected layers), which we vary for both families of the models. In other aspects, the models are similar to those we employed in previous sections - we use batch normalizations with affine transforms for MLP-CNNs, and without for lmKAN-CNNs; MLP-CNNs use ReLU activations, while lmKAN-CNNs do not require additional activation layers. 

The dataset comes with pre-defined full training and test subsets. We split the full training subset into training and validation parts in a 90\%/10\% ratio. Our augmentation pipeline consists of established techniques, such as RandAugment~\cite{cubuk2020randaugment}, MixUp~\cite{zhang2017mixup}, CutMix~\cite{yun2019cutmix}, and a few others. Further details about the augmentations and other aspects of our fitting setup are available in Appendix~\ref{appendix:cifar_10}. 

\begin{figure}[H]
    \centering
    \includegraphics[width=\linewidth]{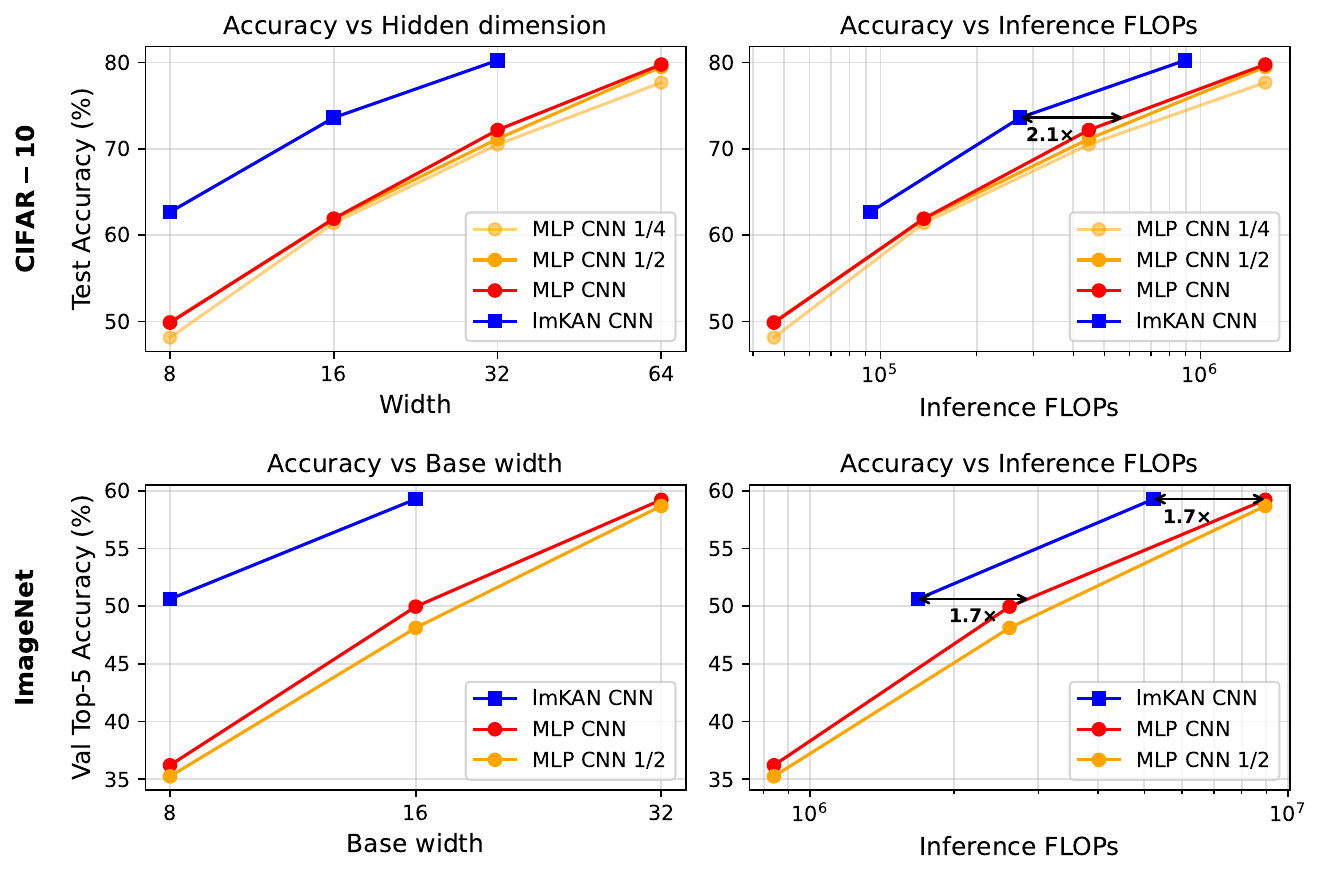}
    \caption{Comparison of the performance of standard MLP-based CNNs and lmKAN-based CNNs on the CIFAR-10 and ImageNet datasets. The "MLP CNN 1/2" line corresponds to the outcome of the fitting procedure with only half of the training steps compared to the "MLP CNN" one.}
    \label{fig:cnn_results}
\end{figure}

Our findings are illustrated in the upper row of Fig.~\ref{fig:cnn_results}. Similarly to previous experiments, lmKAN-based CNNs were found to be more FLOPs efficient compared to classical MLP-based CNNs at the same accuracy level. The observed speed-up factor of 1.6-2.1× is not so dramatic as we report in sections \ref{sec:general_function_approximation} and \ref{sec:methane}, but still substantial. We have not yet implemented dedicated CUDA kernels for efficient inference of lmKAN-based convolutions. Any type of convolution can be cast to a fully connected layer by the corresponding memory manipulations, which we followed during fitting.

\paragraph{Ablations} Appendix~\ref{appendix:cifar_10} contains ablation studies on 1) the effect of the number of grid intervals $G$ and 2) the effect of the strength of Hessian regularization $\lambda$.

\subsubsection{ImageNet} 
To limit the computational cost, we downsampled the images to $81\times81$ ($81 = 3^4$) pixels. Our backbone, illustrated in the right panel of Fig.~\ref{fig:cnn_backbones}, consists of four convolutional layers with the $3\times3$ kernel size and stride $3$ and two fully connected layers. In contrast to the CIFAR-10 experiment, we progressively increase the number of filters as the spatial resolution of the image decreases through the neural network. Our augmentation pipeline consists of the same techniques as the ones we use for CIFAR-10, but with different hyperparameters; see more details in Appendix~\ref{appendix:imagenet}. Since the test subset is not publicly available, we use the validation accuracy as the target metric. The performance of the models is summarized in the bottom row of Fig.~\ref{fig:cnn_results}. The observed efficiency gains of 1.7× are in line with those for CIFAR-10.

\subsection{Comparison with FastKAN}
\label{sec:fastkan_comparison}

We use the training script\footnote{\url{https://github.com/ZiyaoLi/fast-kan/blob/master/examples/train_cifar10.py}} for the CIFAR-10 dataset available in the FastKAN GitHub repository~\cite{fastkan_github} as the basis for the comparison of lmKAN and FastKAN. However, we provide several modifications to the pipeline.

The original script implements the fitting procedure of a fully connected FastKAN model on the CIFAR-10 dataset without augmentations. The model has only one hidden layer with $256$ neurons. Without augmentati ons, it overfits the data quickly. Thus, the very short fitting procedure in the original script is sufficient. 

We extend the script by the same augmentation pipeline as we used in Sec.~\ref{sec:cifar_10} for the CIFAR-10 dataset. We observed that because of augmentations, one has to fit the model for a longer time, so the training budget was substantially increased. Additional changes are detailed in Appendix~\ref{appendix:fastkan}. 

\begin{figure}[h]
    \centering
    \includegraphics[width=0.8\linewidth]{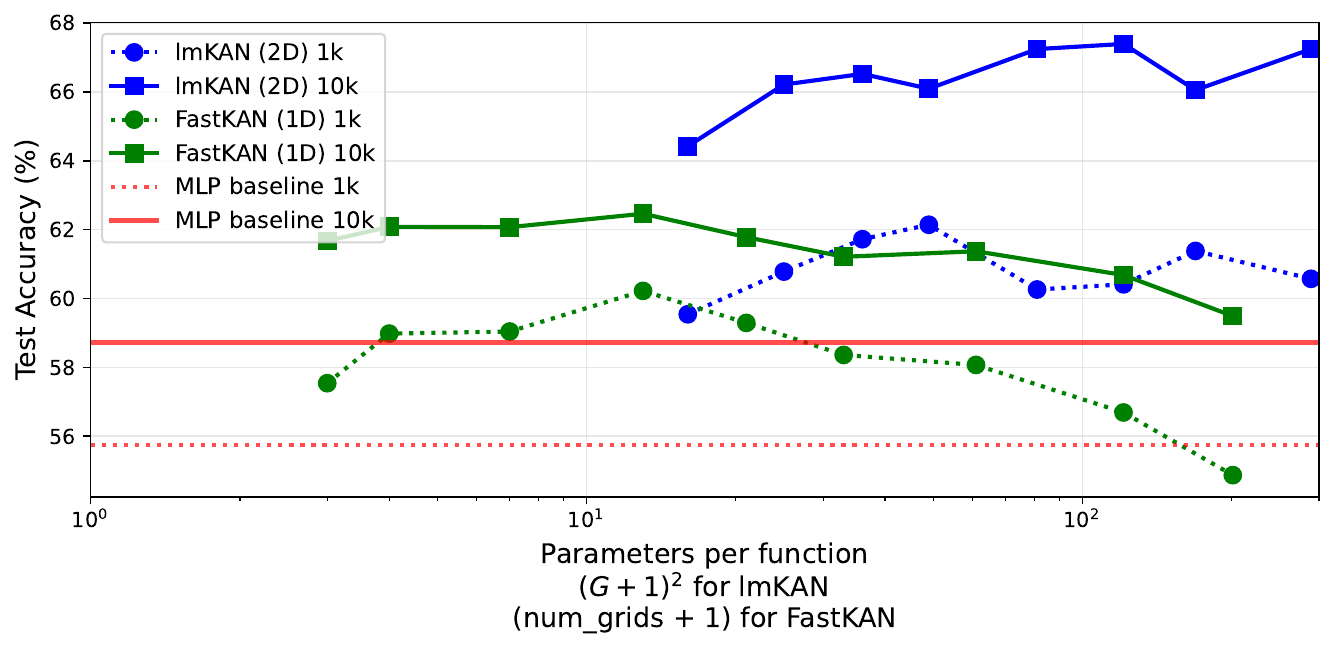}
    \caption{Comparison of lmKAN and FastKAN within the  fully connected framework on the CIFAR-10 dataset. Both models were fit with 1k and 10k epochs.}
    \label{fig:fastkan_comparison_updated}
\end{figure}

These modifications significantly improve the performance of FastKAN models ($54-55\%$ validation accuracy in the original script), see Fig.~\ref{fig:fastkan_comparison_updated}. Furthermore, even the MLP baseline of the same shape yields better accuracy compared to the performance of the FastKAN model in the original script. 

In this section, we systematically compare the performance of the lmKAN and FastKAN models with the same backbone from the original FastKAN script. For each family of the models, we vary the grid resolution and fit the models for one or ten thousand epochs each. The comparison is given in Fig.~\ref{fig:fastkan_comparison_updated}. 

The first observation is that the performance of FastKAN models degrades for excessively fine grid resolutions. For the training budget of one thousand epochs, the final model is even less accurate than the MLP baseline. For the ten thousand epochs, the effect is less pronounced, but it still takes place. For lmKAN models, this degradation is much less severe, if present at all. 

The number of parameters per function scales quadratically with grid resolution for lmKANs and linearly for FastKANs. Thus, although the rightmost lmKAN model in Fig.~\ref{fig:fastkan_comparison_updated} uses more parameters per function, it operates on a much coarser grid --- only $G=16$ grid intervals per dimension, compared with $num\_grids=200$ for FastKANs. This suggests that, for a comparable parameter budget per function, this coarser grid enforces a lower-frequency function class and, in turn, contributes to the superior training stability observed for lmKANs.

Another distinct feature of Fig.~\ref{fig:fastkan_comparison_updated} is that lmKANs achieve notably better accuracy compared to FastKANs, even when the latter are evaluated with a very rich parametrization of inner functions. Additionally, note that the number of trainable functions $[N_{in}/2] N_{out}$ and $N_{in}N_{out}$ in an lmKAN and FastKAN layer, respectively. Thus, the same number of parameters per function corresponds to a two times larger FastKAN model overall. 

To conclude, these findings reinforce the intuitive considerations given in Sec.~\ref{sec:lmkans} and suggest that building blocks of multivariate trainable functions are indeed more effective. 

\section{Limitations}

Many of the difficulties arise when selecting an excessively high number of grid intervals $G$:
\begin{itemize}
    \item When $G$ is too large, lmKANs were found to be hard to converge.
    \item While the throughput does not depend on the grid resolution, both theoretically and empirically, $G$ still affects latency.
    \item Large $G$ entails a large number of parameters; therefore, large models with fine grid resolution have high memory requirements.
\end{itemize}

We have implemented CUDA kernels only for the full precision \texttt{float32} data type. If using data types such as \texttt{bfloat16}, the gains in efficiency are expected similarly to dense matrix multiplications, but such kernels are yet to be implemented. 

\section{Summary}

High-dimensional linear mappings are the cornerstone of modern deep learning, dominating both the parameter count and the computational cost in most models. We introduce lookup multivariate Kolmogorov-Arnold Networks (lmKANs) as a drop-in replacement that offers a substantially better capacity–inference cost trade-off. Across all experiments, lmKANs were Pareto-optimal in the inference FLOPs–performance plane. The efficiency gains are task-dependent: for general high-dimensional function approximation, modeled as a distillation from a large ground-truth teacher MLP with random weights, lmKANs achieved up to 6× fewer FLOPs at matched accuracy. On randomly displaced methane configurations, efficiency improved by up to 78×, or even more, for the {\ttfamily\bfseries Distances} representation. Within convolutional networks, the gains were smaller but still significant: 1.6–2.1× on CIFAR-10 and 1.7× on ImageNet.

Our CUDA kernels compete directly with highly optimized dense matrix multiplications—the backbone of many numerical pipelines for decades. Even so, the gains were sufficient to make lmKANs Pareto-optimal in H100 wall-clock time for both the general high-dimensional function approximation and the methane dataset, achieving the speedup of more than an order of magnitude in the latter case.

\section{Acknowledgments}
S.P. thanks Prashanth Kanduri, Nicholas J. Browning, and Henrique Mendonça for fruitful discussions regarding CUDA, Bob Crovella, whose instrumental lectures are available online\footnote{\url{https://www.youtube.com/playlist?list=PL6RdenZrxrw-zNX7uuGppWETdxt_JxdMj}}, and all the teachers of the CUDA course at Yandex School of Data Analysis. 

S.P. acknowledges support from Intel and Merck KGaA via
the AWASES programme. P.S. acknowledges support from the NCCR Catalysis (grant number
225147), a National Centre of Competence in Research funded by the Swiss National Science Foundation.

\bibliographystyle{unsrtnat}  
\bibliography{references}

\begin{thebibliography}{66}
\providecommand{\natexlab}[1]{#1}
\providecommand{\url}[1]{\texttt{#1}}
\expandafter\ifx\csname urlstyle\endcsname\relax
  \providecommand{\doi}[1]{doi: #1}\else
  \providecommand{\doi}{doi: \begingroup \urlstyle{rm}\Url}\fi

\bibitem[Zhai et~al.(2022)Zhai, Kolesnikov, Houlsby, and Beyer]{zhai2022scaling}
Xiaohua Zhai, Alexander Kolesnikov, Neil Houlsby, and Lucas Beyer.
\newblock Scaling vision transformers.
\newblock In \emph{Proceedings of the IEEE/CVF conference on computer vision and pattern recognition}, pages 12104--12113, 2022.

\bibitem[Kaplan et~al.(2020)Kaplan, McCandlish, Henighan, Brown, Chess, Child, Gray, Radford, Wu, and Amodei]{kaplan2020scaling}
Jared Kaplan, Sam McCandlish, Tom Henighan, Tom~B Brown, Benjamin Chess, Rewon Child, Scott Gray, Alec Radford, Jeffrey Wu, and Dario Amodei.
\newblock Scaling laws for neural language models.
\newblock \emph{arXiv preprint arXiv:2001.08361}, 2020.

\bibitem[Ioffe and Szegedy(2015)]{ioffe2015batch}
Sergey Ioffe and Christian Szegedy.
\newblock Batch normalization: Accelerating deep network training by reducing internal covariate shift.
\newblock In \emph{International conference on machine learning}, pages 448--456. pmlr, 2015.

\bibitem[Hinton et~al.(2012)Hinton, Srivastava, Krizhevsky, Sutskever, and Salakhutdinov]{hinton2012improving}
Geoffrey~E Hinton, Nitish Srivastava, Alex Krizhevsky, Ilya Sutskever, and Ruslan~R Salakhutdinov.
\newblock Improving neural networks by preventing co-adaptation of feature detectors.
\newblock \emph{arXiv preprint arXiv:1207.0580}, 2012.

\bibitem[Vaswani et~al.(2017)Vaswani, Shazeer, Parmar, Uszkoreit, Jones, Gomez, Kaiser, and Polosukhin]{vaswani2017attention}
Ashish Vaswani, Noam Shazeer, Niki Parmar, Jakob Uszkoreit, Llion Jones, Aidan~N Gomez, {\L}ukasz Kaiser, and Illia Polosukhin.
\newblock Attention is all you need.
\newblock \emph{Advances in neural information processing systems}, 30, 2017.

\bibitem[Choromanski et~al.(2020)Choromanski, Likhosherstov, Dohan, Song, Gane, Sarlos, Hawkins, Davis, Mohiuddin, Kaiser, et~al.]{choromanski2020rethinking}
Krzysztof Choromanski, Valerii Likhosherstov, David Dohan, Xingyou Song, Andreea Gane, Tamas Sarlos, Peter Hawkins, Jared Davis, Afroz Mohiuddin, Lukasz Kaiser, et~al.
\newblock Rethinking attention with performers.
\newblock \emph{arXiv preprint arXiv:2009.14794}, 2020.

\bibitem[Elman(1990)]{elman1990finding}
Jeffrey~L Elman.
\newblock Finding structure in time.
\newblock \emph{Cognitive science}, 14\penalty0 (2):\penalty0 179--211, 1990.

\bibitem[Zhou et~al.(2020)Zhou, Cui, Hu, Zhang, Yang, Liu, Wang, Li, and Sun]{zhou2020graph}
Jie Zhou, Ganqu Cui, Shengding Hu, Zhengyan Zhang, Cheng Yang, Zhiyuan Liu, Lifeng Wang, Changcheng Li, and Maosong Sun.
\newblock Graph neural networks: A review of methods and applications.
\newblock \emph{AI open}, 1:\penalty0 57--81, 2020.

\bibitem[LeCun et~al.(2002)LeCun, Bottou, Bengio, and Haffner]{lecun2002gradient}
Yann LeCun, L{\'e}on Bottou, Yoshua Bengio, and Patrick Haffner.
\newblock Gradient-based learning applied to document recognition.
\newblock \emph{Proceedings of the IEEE}, 86\penalty0 (11):\penalty0 2278--2324, 2002.

\bibitem[Krizhevsky et~al.(2012)Krizhevsky, Sutskever, and Hinton]{krizhevsky2012imagenet}
Alex Krizhevsky, Ilya Sutskever, and Geoffrey~E Hinton.
\newblock Imagenet classification with deep convolutional neural networks.
\newblock \emph{Advances in neural information processing systems}, 25, 2012.

\bibitem[Liu et~al.(2024)Liu, Wang, Vaidya, Ruehle, Halverson, Solja{\v{c}}i{\'c}, Hou, and Tegmark]{liu2024kan}
Ziming Liu, Yixuan Wang, Sachin Vaidya, Fabian Ruehle, James Halverson, Marin Solja{\v{c}}i{\'c}, Thomas~Y Hou, and Max Tegmark.
\newblock Kan: Kolmogorov-arnold networks.
\newblock \emph{arXiv preprint arXiv:2404.19756}, 2024.

\bibitem[Li(2024{\natexlab{a}})]{li2024kolmogorovarnold}
Ziyao Li.
\newblock Kolmogorov-arnold networks are radial basis function networks.
\newblock 2024{\natexlab{a}}.

\bibitem[Kolmogorov()]{kolmogorov1961representation}
Andre{\u\i} Kolmogorov.
\newblock \emph{On the representation of continuous functions of several variables by superpositions of continuous functions of a smaller number of variables}.

\bibitem[Arnold(2009)]{arnold2009functions}
Vladimir~I Arnold.
\newblock On functions of three variables.
\newblock \emph{Collected Works: Representations of Functions, Celestial Mechanics and KAM Theory, 1957--1965}, pages 5--8, 2009.

\bibitem[Girosi and Poggio(1989)]{girosi1989representation}
Federico Girosi and Tomaso Poggio.
\newblock Representation properties of networks: Kolmogorov's theorem is irrelevant.
\newblock \emph{Neural Computation}, 1\penalty0 (4):\penalty0 465--469, 1989.

\bibitem[Schmidt-Hieber(2021)]{schmidt2021kolmogorov}
Johannes Schmidt-Hieber.
\newblock The kolmogorov--arnold representation theorem revisited.
\newblock \emph{Neural networks}, 137:\penalty0 119--126, 2021.

\bibitem[Hecht-Nielsen(1987)]{hecht1987kolmogorov}
Robert Hecht-Nielsen.
\newblock Kolmogorov’s mapping neural network existence theorem.
\newblock In \emph{Proceedings of the international conference on Neural Networks}, volume~3, pages 11--14. IEEE press New York, NY, USA, 1987.

\bibitem[Igelnik and Parikh(2003)]{igelnik2003kolmogorov}
Boris Igelnik and Neel Parikh.
\newblock Kolmogorov's spline network.
\newblock \emph{IEEE transactions on neural networks}, 14\penalty0 (4):\penalty0 725--733, 2003.

\bibitem[Yang and Wang(2024)]{yang2024kolmogorov}
Xingyi Yang and Xinchao Wang.
\newblock Kolmogorov-arnold transformer.
\newblock \emph{arXiv preprint arXiv:2409.10594}, 2024.

\bibitem[Kundu et~al.(2024)Kundu, Sarkar, and Sadhu]{kundu2024kanqas}
Akash Kundu, Aritra Sarkar, and Abhishek Sadhu.
\newblock Kanqas: Kolmogorov-arnold network for quantum architecture search.
\newblock \emph{EPJ Quantum Technology}, 11\penalty0 (1):\penalty0 76, 2024.

\bibitem[Kashefi(2025)]{kashefi2025pointnet}
Ali Kashefi.
\newblock Pointnet with kan versus pointnet with mlp for 3d classification and segmentation of point sets.
\newblock \emph{Computers \& Graphics}, page 104319, 2025.

\bibitem[Yu et~al.(2024)Yu, Yu, and Wang]{yu2024kan}
Runpeng Yu, Weihao Yu, and Xinchao Wang.
\newblock Kan or mlp: A fairer comparison.
\newblock \emph{arXiv preprint arXiv:2407.16674}, 2024.

\bibitem[Somvanshi et~al.(2024)Somvanshi, Javed, Islam, Pandit, and Das]{somvanshi2024survey}
Shriyank Somvanshi, Syed~Aaqib Javed, Md~Monzurul Islam, Diwas Pandit, and Subasish Das.
\newblock A survey on kolmogorov-arnold network.
\newblock \emph{ACM Computing Surveys}, 2024.

\bibitem[Ji et~al.(2024)Ji, Hou, and Zhang]{ji2024comprehensive}
Tianrui Ji, Yuntian Hou, and Di~Zhang.
\newblock A comprehensive survey on kolmogorov arnold networks (kan).
\newblock \emph{arXiv preprint arXiv:2407.11075}, 2024.

\bibitem[SS et~al.(2024)SS, AR, KP, et~al.]{ss2024chebyshev}
Sidharth SS, Keerthana AR, Anas KP, et~al.
\newblock Chebyshev polynomial-based kolmogorov-arnold networks: An efficient architecture for nonlinear function approximation.
\newblock \emph{arXiv preprint arXiv:2405.07200}, 2024.

\bibitem[Xu et~al.(2024)Xu, Chen, Li, Yang, Wang, Hu, and Ngai]{xu2024fourierkan}
Jinfeng Xu, Zheyu Chen, Jinze Li, Shuo Yang, Wei Wang, Xiping Hu, and Edith C-H Ngai.
\newblock Fourierkan-gcf: Fourier kolmogorov-arnold network--an effective and efficient feature transformation for graph collaborative filtering.
\newblock \emph{arXiv preprint arXiv:2406.01034}, 2024.

\bibitem[Moradzadeh et~al.(2024)Moradzadeh, Wawrzyniak, Macklin, and Paliwal]{moradzadeh2024ukan}
Alireza Moradzadeh, Lukasz Wawrzyniak, Miles Macklin, and Saee~G Paliwal.
\newblock Ukan: Unbound kolmogorov-arnold network accompanied with accelerated library.
\newblock \emph{arXiv preprint arXiv:2408.11200}, 2024.

\bibitem[Huang et~al.(2025)Huang, Jia, Kong, Waqar, Wen, Chang, and Yu]{huang2025hardware}
Wei-Hsing Huang, Jianwei Jia, Yuyao Kong, Faaiq Waqar, Tai-Hao Wen, Meng-Fan Chang, and Shimeng Yu.
\newblock Hardware acceleration of kolmogorov-arnold network (kan) for lightweight edge inference.
\newblock In \emph{Proceedings of the 30th Asia and South Pacific Design Automation Conference}, pages 693--699, 2025.

\bibitem[Poluektov and Polar(2025)]{poluektov2025construction}
Michael Poluektov and Andrew Polar.
\newblock Construction of the kolmogorov-arnold networks using the newton-kaczmarz method.
\newblock \emph{Machine Learning}, 114\penalty0 (8):\penalty0 185, 2025.

\bibitem[Polar and Poluektov(2021)]{polar2021deep}
Andrew Polar and Michael Poluektov.
\newblock A deep machine learning algorithm for construction of the kolmogorov--arnold representation.
\newblock \emph{Engineering Applications of Artificial Intelligence}, 99:\penalty0 104137, 2021.

\bibitem[He et~al.(2016)He, Zhang, Ren, and Sun]{he2016deep}
Kaiming He, Xiangyu Zhang, Shaoqing Ren, and Jian Sun.
\newblock Deep residual learning for image recognition.
\newblock In \emph{Proceedings of the IEEE conference on computer vision and pattern recognition}, pages 770--778, 2016.

\bibitem[De~Boor and De~Boor(1978)]{de1978practical}
Carl De~Boor and Carl De~Boor.
\newblock \emph{A practical guide to splines}, volume~27.
\newblock springer New York, 1978.

\bibitem[De~Boor(1968)]{de1968uniform}
Carl De~Boor.
\newblock On uniform approximation by splines.
\newblock \emph{J. Approx. Theory}, 1\penalty0 (1):\penalty0 219--235, 1968.

\bibitem[Glorot et~al.(2011)Glorot, Bordes, and Bengio]{glorot2011deep}
Xavier Glorot, Antoine Bordes, and Yoshua Bengio.
\newblock Deep sparse rectifier neural networks.
\newblock In \emph{Proceedings of the fourteenth international conference on artificial intelligence and statistics}, pages 315--323. JMLR Workshop and Conference Proceedings, 2011.

\bibitem[Volkov and Demmel(2008)]{volkov2008benchmarking}
Vasily Volkov and James~W Demmel.
\newblock Benchmarking gpus to tune dense linear algebra.
\newblock In \emph{SC'08: Proceedings of the 2008 ACM/IEEE conference on Supercomputing}, pages 1--11. IEEE, 2008.

\bibitem[Xie et~al.(2023)Xie, Rupp, and Hennig]{xie2023ultra}
Stephen~R Xie, Matthias Rupp, and Richard~G Hennig.
\newblock Ultra-fast interpretable machine-learning potentials.
\newblock \emph{npj Computational Materials}, 9\penalty0 (1):\penalty0 162, 2023.

\bibitem[Srivastava et~al.(2014)Srivastava, Hinton, Krizhevsky, Sutskever, and Salakhutdinov]{srivastava2014dropout}
Nitish Srivastava, Geoffrey Hinton, Alex Krizhevsky, Ilya Sutskever, and Ruslan Salakhutdinov.
\newblock Dropout: a simple way to prevent neural networks from overfitting.
\newblock \emph{The journal of machine learning research}, 15\penalty0 (1):\penalty0 1929--1958, 2014.

\bibitem[Cybenko(1989)]{cybenko1989approximation}
George Cybenko.
\newblock Approximation by superpositions of a sigmoidal function.
\newblock \emph{Mathematics of control, signals and systems}, 2\penalty0 (4):\penalty0 303--314, 1989.

\bibitem[Kingma and Ba(2014)]{kingma2014adam}
Diederik~P Kingma and Jimmy Ba.
\newblock Adam: A method for stochastic optimization.
\newblock \emph{arXiv preprint arXiv:1412.6980}, 2014.

\bibitem[{Kaggle}()]{KaggleTitanic}
{Kaggle}.
\newblock Titanic — machine learning from disaster.
\newblock \url{https://www.kaggle.com/competitions/titanic}.
\newblock Accessed 2025-08-19.

\bibitem[Harrison~Jr and Rubinfeld(1978)]{harrison1978hedonic}
David Harrison~Jr and Daniel~L Rubinfeld.
\newblock Hedonic housing prices and the demand for clean air.
\newblock \emph{Journal of environmental economics and management}, 5\penalty0 (1):\penalty0 81--102, 1978.

\bibitem[Pozdnyakov et~al.(2020)Pozdnyakov, Willatt, Bart{\'o}k, Ortner, Cs{\'a}nyi, and Ceriotti]{pozdnyakov2020incompleteness}
Sergey~N Pozdnyakov, Michael~J Willatt, Albert~P Bart{\'o}k, Christoph Ortner, G{\'a}bor Cs{\'a}nyi, and Michele Ceriotti.
\newblock Incompleteness of atomic structure representations.
\newblock \emph{Physical Review Letters}, 125\penalty0 (16):\penalty0 166001, 2020.

\bibitem[Turney et~al.(2012)Turney, Simmonett, Parrish, Hohenstein, Evangelista, Fermann, Mintz, Burns, Wilke, Abrams, et~al.]{turney2012psi4}
Justin~M Turney, Andrew~C Simmonett, Robert~M Parrish, Edward~G Hohenstein, Francesco~A Evangelista, Justin~T Fermann, Benjamin~J Mintz, Lori~A Burns, Jeremiah~J Wilke, Micah~L Abrams, et~al.
\newblock Psi4: an open-source ab initio electronic structure program.
\newblock \emph{Wiley Interdisciplinary Reviews: Computational Molecular Science}, 2\penalty0 (4):\penalty0 556--565, 2012.

\bibitem[Kohn and Sham(1965)]{kohn1965self}
Walter Kohn and Lu~Jeu Sham.
\newblock Self-consistent equations including exchange and correlation effects.
\newblock \emph{Physical review}, 140\penalty0 (4A):\penalty0 A1133, 1965.

\bibitem[Behler and Parrinello(2007)]{behler2007generalized}
J{\"o}rg Behler and Michele Parrinello.
\newblock Generalized neural-network representation of high-dimensional potential-energy surfaces.
\newblock \emph{Physical review letters}, 98\penalty0 (14):\penalty0 146401, 2007.

\bibitem[Bart{\'o}k et~al.(2010)Bart{\'o}k, Payne, Kondor, and Cs{\'a}nyi]{bartok2010gaussian}
Albert~P Bart{\'o}k, Mike~C Payne, Risi Kondor, and G{\'a}bor Cs{\'a}nyi.
\newblock Gaussian approximation potentials: The accuracy of quantum mechanics, without the electrons.
\newblock \emph{Physical review letters}, 104\penalty0 (13):\penalty0 136403, 2010.

\bibitem[Allen et~al.(2021)Allen, Dusson, Ortner, and Cs{\'a}nyi]{allen2021atomic}
Alice~EA Allen, Genevi{\`e}ve Dusson, Christoph Ortner, and G{\'a}bor Cs{\'a}nyi.
\newblock Atomic permutationally invariant polynomials for fitting molecular force fields.
\newblock \emph{Machine Learning: Science and Technology}, 2\penalty0 (2):\penalty0 025017, 2021.

\bibitem[Zhang et~al.(2021)Zhang, Xia, and Jiang]{zhang2021physically}
Yaolong Zhang, Junfan Xia, and Bin Jiang.
\newblock Physically motivated recursively embedded atom neural networks: incorporating local completeness and nonlocality.
\newblock \emph{Physical Review Letters}, 127\penalty0 (15):\penalty0 156002, 2021.

\bibitem[Pozdnyakov and Ceriotti(2023)]{pozdnyakov2023smooth}
Sergey Pozdnyakov and Michele Ceriotti.
\newblock Smooth, exact rotational symmetrization for deep learning on point clouds.
\newblock \emph{Advances in Neural Information Processing Systems}, 36:\penalty0 79469--79501, 2023.

\bibitem[Bodner et~al.(2024)Bodner, Tepsich, Spolski, and Pourteau]{bodner2024convolutional}
Alexander~Dylan Bodner, Antonio~Santiago Tepsich, Jack~Natan Spolski, and Santiago Pourteau.
\newblock Convolutional kolmogorov-arnold networks.
\newblock \emph{arXiv preprint arXiv:2406.13155}, 2024.

\bibitem[Krizhevsky et~al.(2009)Krizhevsky, Hinton, et~al.]{krizhevsky2009learning}
Alex Krizhevsky, Geoffrey Hinton, et~al.
\newblock Learning multiple layers of features from tiny images.
\newblock 2009.

\bibitem[Deng et~al.(2009)Deng, Dong, Socher, Li, Li, and Fei-Fei]{deng2009imagenet}
Jia Deng, Wei Dong, Richard Socher, Li-Jia Li, Kai Li, and Li~Fei-Fei.
\newblock Imagenet: A large-scale hierarchical image database.
\newblock In \emph{2009 IEEE conference on computer vision and pattern recognition}, pages 248--255. Ieee, 2009.

\bibitem[Cubuk et~al.(2020)Cubuk, Zoph, Shlens, and Le]{cubuk2020randaugment}
Ekin~D Cubuk, Barret Zoph, Jonathon Shlens, and Quoc~V Le.
\newblock Randaugment: Practical automated data augmentation with a reduced search space.
\newblock In \emph{Proceedings of the IEEE/CVF conference on computer vision and pattern recognition workshops}, pages 702--703, 2020.

\bibitem[Zhang et~al.(2017)Zhang, Cisse, Dauphin, and Lopez-Paz]{zhang2017mixup}
Hongyi Zhang, Moustapha Cisse, Yann~N Dauphin, and David Lopez-Paz.
\newblock mixup: Beyond empirical risk minimization.
\newblock \emph{arXiv preprint arXiv:1710.09412}, 2017.

\bibitem[Yun et~al.(2019)Yun, Han, Oh, Chun, Choe, and Yoo]{yun2019cutmix}
Sangdoo Yun, Dongyoon Han, Seong~Joon Oh, Sanghyuk Chun, Junsuk Choe, and Youngjoon Yoo.
\newblock Cutmix: Regularization strategy to train strong classifiers with localizable features.
\newblock In \emph{Proceedings of the IEEE/CVF international conference on computer vision}, pages 6023--6032, 2019.

\bibitem[Li(2024{\natexlab{b}})]{fastkan_github}
Ziyao Li.
\newblock {fast-kan}: Fastkan — very fast implementation of kolmogorov–arnold networks (kan).
\newblock \url{https://github.com/ZiyaoLi/fast-kan}, 2024{\natexlab{b}}.
\newblock GitHub repository.

\bibitem[Bengio et~al.(2013)Bengio, L{\'e}onard, and Courville]{bengio2013estimating}
Yoshua Bengio, Nicholas L{\'e}onard, and Aaron Courville.
\newblock Estimating or propagating gradients through stochastic neurons for conditional computation.
\newblock \emph{arXiv preprint arXiv:1308.3432}, 2013.

\bibitem[Bengio et~al.(2015)Bengio, Bacon, Pineau, and Precup]{bengio2015conditional}
Emmanuel Bengio, Pierre-Luc Bacon, Joelle Pineau, and Doina Precup.
\newblock Conditional computation in neural networks for faster models.
\newblock \emph{arXiv preprint arXiv:1511.06297}, 2015.

\bibitem[Jacobs et~al.(1991)Jacobs, Jordan, Nowlan, and Hinton]{jacobs1991adaptive}
Robert~A Jacobs, Michael~I Jordan, Steven~J Nowlan, and Geoffrey~E Hinton.
\newblock Adaptive mixtures of local experts.
\newblock \emph{Neural computation}, 3\penalty0 (1):\penalty0 79--87, 1991.

\bibitem[Shazeer et~al.(2017)Shazeer, Mirhoseini, Maziarz, Davis, Le, Hinton, and Dean]{shazeer2017outrageously}
Noam Shazeer, Azalia Mirhoseini, Krzysztof Maziarz, Andy Davis, Quoc Le, Geoffrey Hinton, and Jeff Dean.
\newblock Outrageously large neural networks: The sparsely-gated mixture-of-experts layer.
\newblock \emph{arXiv preprint arXiv:1701.06538}, 2017.

\bibitem[Fedus et~al.(2022)Fedus, Zoph, and Shazeer]{fedus2022switch}
William Fedus, Barret Zoph, and Noam Shazeer.
\newblock Switch transformers: Scaling to trillion parameter models with simple and efficient sparsity.
\newblock \emph{Journal of Machine Learning Research}, 23\penalty0 (120):\penalty0 1--39, 2022.

\bibitem[Wang et~al.(2018)Wang, Suo, Ma, Pokrovsky, and Urtasun]{wang2018deep}
Shenlong Wang, Simon Suo, Wei-Chiu Ma, Andrei Pokrovsky, and Raquel Urtasun.
\newblock Deep parametric continuous convolutional neural networks.
\newblock In \emph{Proceedings of the IEEE conference on computer vision and pattern recognition}, pages 2589--2597, 2018.

\bibitem[Batatia et~al.(2022)Batatia, Kovacs, Simm, Ortner, and Cs{\'a}nyi]{batatia2022mace}
Ilyes Batatia, David~P Kovacs, Gregor Simm, Christoph Ortner, and G{\'a}bor Cs{\'a}nyi.
\newblock Mace: Higher order equivariant message passing neural networks for fast and accurate force fields.
\newblock \emph{Advances in neural information processing systems}, 35:\penalty0 11423--11436, 2022.

\bibitem[Pozdnyakov et~al.(2023)Pozdnyakov, Oganov, Mazhnik, Mazitov, and Kruglov]{pozdnyakov2023fast}
Sergey Pozdnyakov, Artem~R Oganov, Efim Mazhnik, Arslan Mazitov, and Ivan Kruglov.
\newblock Fast general two-and three-body interatomic potential.
\newblock \emph{Physical Review B}, 107\penalty0 (12):\penalty0 125160, 2023.

\bibitem[Derksen and Kemper(2015)]{derksen2015computational}
Harm Derksen and Gregor Kemper.
\newblock \emph{Computational invariant theory}.
\newblock Springer, 2015.

\bibitem[Loshchilov and Hutter(2016)]{loshchilov2016sgdr}
Ilya Loshchilov and Frank Hutter.
\newblock Sgdr: Stochastic gradient descent with warm restarts.
\newblock \emph{arXiv preprint arXiv:1608.03983}, 2016.

\end{thebibliography}

\newpage

\appendix 

\section{Additional related work}
\label{appendix:additional_related_work}
In this appendix, we review further related work beyond Kolmogorov-Arnold Networks. 

\paragraph{Conditional computations} The proposed lookup multivariate Kolmogorov-Arnold Networks fit within the general idea of conditional computation~\cite{bengio2013estimating,bengio2015conditional}, which suggests using only part of a model's parameters at inference time. The most popular subclass of such architectures is the Mixture-of-Experts (MoE) family~\cite{jacobs1991adaptive, shazeer2017outrageously, fedus2022switch}, in which multiple experts, typically implemented as MLPs, process the current input only if they are deemed relevant.

In the proposed method, the spline lookup tables enable $\mathcal{O}(1)$ computations, which underpin the observed efficiency of our approach. For a function parametrized in this way, its value at any given point depends only on the small subset of parameters that control the function's behavior within the corresponding grid interval; all other parameters remain inactive, similarly to other conditional-computation methods.

Whereas our method can be loosely characterized as performing low-level "weight lookup," classical MoE models can be viewed as performing high-level "expert lookup." In these models, entire experts—typically full MLPs—or at least entire layers are either active or inactive as a whole. In general, such high-level approaches dominate the landscape of conditional-computation models, while low-level ones are very rare, likely due to the challenges associated with efficient GPU implementation.

\paragraph{Spline lookup tables} Outside the scope of general high-dimensional mappings, spline lookup tables have found several applications in machine learning. For instance, Graph Neural Networks (GNNs) applied to point clouds in Euclidean space~\cite{wang2018deep} often have trainable filters as part of the model. Such filters are one- or three-dimensional functions whose inputs are inter-point distances or displacement vectors. Sometimes these functions are splined~\cite{batatia2022mace} post-hoc, after the fitting of the model is finished. 

In a more niche case, certain physics-inspired machine learning models can be parametrized \emph{exclusively} by low-dimensional functions. For such setups, \citet{pozdnyakov2023fast} and \citet{xie2023ultra} pushed the boundary of efficiency by applying B-spline techniques similarly to this work.

\section{More details on function parametrization}
\label{appendix:func_parametrization}
\paragraph{The $\sigma(x)$ function}

Section~\ref{sec:functions_parametrization} of the main text describes the construction of the static percentile grid used in the parametrization of all the functions lmKAN consists of. This construction involves any sigmoid-like function $\sigma(x)$, and, as was discussed in the main text, it makes sense to shape it reasonably close to cumulative normal distribution to distribute function arguments $x$ evenly across grid intervals. 

\begin{figure}[h]%
    \centering
    \begin{minipage}[c]{0.52\linewidth}
        \vspace{-3.3\baselineskip}
        \begin{equation}\label{eq:sigma}
            \sigma(x)=
            \begin{cases}
                0.5\,e^{x}, & x\le 0,\\[6pt]
                1-0.5\,e^{-x}, & x>0.
            \end{cases}
        \end{equation}
    \end{minipage}\hfill
    \begin{minipage}[c]{0.44\linewidth}
        \centering
        \includegraphics[width=\linewidth]{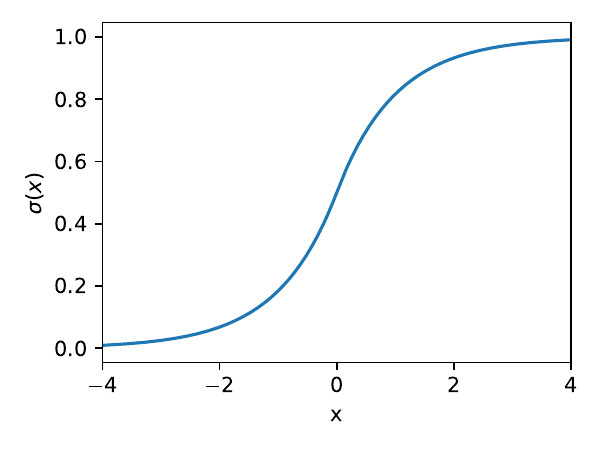}
        \caption{Plot of the $\sigma(x)$ function defined in Eq.~\ref{eq:sigma}.}
        \label{fig:sigma_func}
    \end{minipage}
\end{figure}

It is, however, computationally expensive to use the exact cumulative distribution function of the normal distribution. Thus, we use a fast approximation, which is given in Eq.~\ref{eq:sigma} and is illustrated in Fig.~\ref{fig:sigma_func}. 

This construction is cheap to compute because the computational pipeline consists of a single exponential call and a few arithmetic operations, as elaborated in algorithm~\ref{algorithm:sigma}. 

\begin{algorithm}[H]
\caption{Evaluation of $\sigma(x)$ with a single exponential call}
\label{alg:sigma}
\begin{algorithmic}[1]
\Require $x \in \mathbb{R}$
\Ensure $\sigma(x)$
\State $t \gets \exp\!\bigl(-|x|\bigr)$  \Comment{compute expensive exponential only once}
\If{$x > 0$}
    \State $\sigma(x) \gets 1 - 0.5 \, t$
\Else
    \State $\sigma(x) \gets 0.5 \, t$
\EndIf
\State \Return $\sigma(x)$
\end{algorithmic}
\label{algorithm:sigma}
\end{algorithm}

\paragraph{Edge cases}
Section~\ref{sec:functions_parametrization} of the main text introduced static percentile grids and the corresponding basis of second-order B-splines. For a grid with $G$ intervals and $G - 1$ grid points, there are $G + 1$ basis functions, out of which $G - 1$ are given by second-order B-splines centered around all grid points, as illustrated in the right panel of Fig.~\ref{fig:cdf_grid} of the main text. The other two are given as linear functions on the left-most and right-most infinite intervals. 

The second-order B-splines given in the right panel of Fig.~\ref{fig:cdf_grid} are defined to linearly increase from $0$ to $1$ from the left grid point to the central one and then linearly decrease from $1$ to $0$ from the central grid point to the right grid point. This construction is well-defined for all the inner grid points but requires additional definitions for the B-splines centered around the left-most and right-most grid points, as these do not have left and right neighboring grid points, respectively. 

In order to define these edge B-splines, we introduce 'ghost' left and right grid points. The position of the left ghost point is given as $\bm{\mathcal{G}}[0] - (\bm{\mathcal{G}}[1] - \bm{\mathcal{G}}[0])$, where $\bm{\mathcal{G}}[0]$ and $\bm{\mathcal{G}}[1]$ are the positions of left-most and second left-most grid points respectively. The right ghost point is defined similarly. With such a notation of additional grid points, we can define edge B-splines similarly to all the others. 

Finally, there are two linear basis functions on the left-most and right-most infinite intervals. We define them to be zeros in the left-most and right-most grid points, linearly increasing to ones at the left and right ghost points and continuing to left and right infinities with the same slope, respectively.

\paragraph{Difference between the direct use of the static percentile grid and the uniform one with pre-normalization by $\sigma(x)$} 

A reader may ask themselves a question of what is the difference between the proposed construction involving a function $f(x)$ defined on static percentile grid on $(-\infty, \infty)$ and a more simple approach which involves first mapping of an input to $(0, 1)$ by $x' = \sigma(x)$, and then computing a piecewise linear function $g(x')$ defined on a uniform grid on $[0, 1]$. 

The answer is that the resulting functions in the initial space, $f(x)$ and $f'(x) = g(\sigma(x))$, are not identical. Specifically, $f'(x)$ is no longer piecewise linear on each grid interval, with the most notable difference being the behavior when $x \rightarrow \pm \infty$. Asymptotes of $f(x)$ are linear, while asymptotes of $f'(x)$ are constant. 

The concern with horizontal asymptotes is that they can cause the vanishing gradient problem, which is believed to be one of the reasons why the ReLU~\cite{glorot2011deep} activations work better than \texttt{tanh}. Given these considerations, we chose the direct application of the static percentile grid for practical implementation.

\subsection{Standalone two-dimensional function computation}
\label{appendix:2d_function_computation}
Algorithm~\ref{algorithm:lmKAN2D} provides a recipe to compute a standalone two-dimensional function given our parametrization scheme. See discussion in Sec.~\ref{sec:functions_parametrization} of the main text. 

\begin{algorithm}[th]
  \caption{$\mathcal{O}(1)$ evaluation of a standalone 2D lmKAN function. Red lines indicate computations that can be reused when computing many 2D functions for the same arguments, while green lines indicate computations that have to be repeated for each 2D function.}
  \label{algorithm:lmKAN2D}
  \begin{algorithmic}[1]
    \Require scalars $x_1,x_2\in\mathbb{R}$; grid points $\bm{\mathcal{G}}$;
            parameters $\mathbf{P}\in\mathbb{R}^{[G+1,G+1]}$ \newline
            $\mathbf{P}[\,i_1,\,i_2\,]$ stores the function value on the $(i_1,i_2)$-th grid point
    \Ensure output $y\in\mathbb{R}$

    \Statex
    \Function{\textbf{Eval2D}}{$x_1,x_2$}
      \Comment{Handling of edge-interval cases described above is omitted for brevity.}
      \State {\color{darkred}$i_1 \gets \bigl\lfloor \sigma(x_1)\,G \bigr\rfloor$}
      \State {\color{darkred}$i_2 \gets \bigl\lfloor \sigma(x_2)\,G \bigr\rfloor$}
       \State {\color{darkred}$B_{i_1\,i_2}(x_1,x_2)
      \gets
      \dfrac{\bm{\mathcal{G}}[i_1+1]-x_1}
            {\bm{\mathcal{G}}[i_1+1]-\bm{\mathcal{G}}[i_1]}
      \;
      \dfrac{\bm{\mathcal{G}}[i_2+1]-x_2}
            {\bm{\mathcal{G}}[i_2+1]-\bm{\mathcal{G}}[i_2]}$}
      \State {\color{darkred}$B_{i_1+1\,i_2}(x_1,x_2)
      \gets
      \dfrac{x_1-\bm{\mathcal{G}}[i_1]}
            {\bm{\mathcal{G}}[i_1+1]-\bm{\mathcal{G}}[i_1]}
      \;
      \dfrac{\bm{\mathcal{G}}[i_2+1]-x_2}
            {\bm{\mathcal{G}}[i_2+1]-\bm{\mathcal{G}}[i_2]}$}
      \State {\color{darkred}$B_{i_1\,i_2+1}(x_1,x_2)
      \gets
      \dfrac{\bm{\mathcal{G}}[i_1+1]-x_1}
            {\bm{\mathcal{G}}[i_1+1]-\bm{\mathcal{G}}[i_1]}
      \;
      \dfrac{x_2-\bm{\mathcal{G}}[i_2]}
            {\bm{\mathcal{G}}[i_2+1]-\bm{\mathcal{G}}[i_2]}$}
      \State {\color{darkred}$B_{i_1+1\,i_2+1}(x_1,x_2)
      \gets
      \dfrac{x_1-\bm{\mathcal{G}}[i_1]}
            {\bm{\mathcal{G}}[i_1+1]-\bm{\mathcal{G}}[i_1]}
      \;
      \dfrac{x_2-\bm{\mathcal{G}}[i_2]}
            {\bm{\mathcal{G}}[i_2+1]-\bm{\mathcal{G}}[i_2]}$}
      \State $y \gets 0$
      \State {\color{darkgreen}$y \mathrel{+}= B_{i_1\,i_2}(x_1,x_2)\;\mathbf{P}[\,i_1,\,i_2\,]$}
      \State {\color{darkgreen}$y \mathrel{+}= B_{i_1+1\,i_2}(x_1,x_2)\;\mathbf{P}[\,i_1\!+\!1,\,i_2\,]$}
      \State {\color{darkgreen}$y \mathrel{+}= B_{i_1\,i_2+1}(x_1,x_2)\;\mathbf{P}[\,i_1,\,i_2\!+\!1\,]$}
      \State {\color{darkgreen}$y \mathrel{+}= B_{i_1+1\,i_2+1}(x_1,x_2)\;\mathbf{P}[\,i_1\!+\!1,\,i_2\!+\!1\,]$}
      \State \Return $y$
    \EndFunction
  \end{algorithmic}
\end{algorithm}

\section{More details on Hessian regularization}
\label{appendix:off_diagonal_regularization}
Sec.~\ref{sec:hessian_regularization} of the main text introduced the concept of off-diagonal Hessian regularization which is based on finite difference schemes for squared Frobenius norm of the Hessian. This appendix provides the exact equations we use in our implementation. 

We use the following finite-differences approximation for the second derivative with respect to $x_1$:

\begin{equation}
    \left.\frac{\partial^{2} f}{\partial x_{1}^{2}}\right|_{(x_{1},x_{2})}
\;\approx\;
\frac{2\bigl(
      h_{\ell}\,f(x_{1}+h_{r},\,x_{2})
      - (h_{\ell}+h_{r})\,f(x_{1},\,x_{2})
      + h_{r}\,f(x_{1}-h_{\ell},\,x_{2})
      \bigr)}
     {h_{\ell}\,h_{r}\,(h_{\ell}+h_{r})}\text{,}
\end{equation}
where $h_l$ is the spacing between left and central grid points, while $h_r$ is the spacing between central and right grid points. 
The corresponding expression in terms of the coefficients $p_{i, j}$ is given as:

\begin{equation}
    D_{x_1, x_1; i, j} = 
\frac{2\Bigl(
      h_{i}\,p_{i+1,j}
      \;-\;(h_{i}+h_{i+1})\,p_{i,j}
      \;+\;h_{i+1}\,p_{i-1,j}
      \Bigr)}
     {h_{i}\,h_{i+1}\,(h_{i}+h_{i+1})}
\end{equation}

For the second derivative with respect to $x_2$ we use an analogous expression:

\begin{equation}
    \left.\frac{\partial^{2} f}{\partial x_{2}^{2}}\right|_{(x_{1},x_{2})}
\;\approx\;
\frac{2\Bigl(
      h_{b}\,f(x_{1},\,x_{2}+h_{u})
      - (h_{b}+h_{u})\,f(x_{1},\,x_{2})
      + h_{u}\,f(x_{1},\,x_{2}-h_{b})
      \Bigr)}
     {h_{b}\,h_{u}\,(h_{b}+h_{u})}\text{,}
\end{equation}
where $h_u$ and $h_b$ are upper and bottom spacings, respectively. 
\begin{equation}
D_{x_2, x_2; i, j} =
\frac{2\Bigl(
      h_{j}\,p_{i,j+1}
      \;-\;(h_{j}+h_{j+1})\,p_{i,j}
      \;+\;h_{j+1}\,p_{i,j-1}
      \Bigr)}
     {h_{j}\,h_{j+1}\,(h_{j}+h_{j+1})}
\end{equation}

For the mixed derivative, our finite-differences scheme is the following:
\begin{equation}
\begin{aligned}
&\left.\frac{\partial^{2} f}{\partial x_{1}\,\partial x_{2}}\right|_{(x_{1},x_{2})}
\;\approx \\[6pt]
&\frac{
      f(x_{1}+h_{r},\,x_{2}+h_{p})
      - f(x_{1}+h_{r},\,x_{2}-h_{b})
      - f(x_{1}-h_{l},\,x_{2}+h_{p})
      + f(x_{1}-h_{l},\,x_{2}-h_{b})
     }
     {(h_{r}+h_{l})\,(h_{p}+h_{b})}
\end{aligned}
\end{equation}

\begin{equation}
    \begin{aligned}
D_{x_1, x_2; i, j} = 
&\frac{
      p_{i+1,j+1}
      - p_{i+1,j-1}
      - p_{i-1,j+1}
      + p_{i-1,j-1}
     }
     {(h_{i}+h_{i+1})\,(h_{j}+h_{j+1})}
\end{aligned}
\end{equation}

The final regularization term is the following:

\begin{equation}
    H_{i, j} = D_{x_1, x_1; i, j}^2 + 2D_{x_1, x_2; i, j}^2 + D_{x_2, x_2; i, j}^2\text{.}
\end{equation}

In order to compute the total regularization term for the whole model, we 1) average $H_{i, j}$ across all the grid points and 2) summate these values across all the 2D functions within all the lmKAN layers in the model. 

\section{CUDA kernels}
\label{appendix:cuda_kernels}
\begin{figure}[h]
    \centering
    \includegraphics[width=1.0\linewidth]{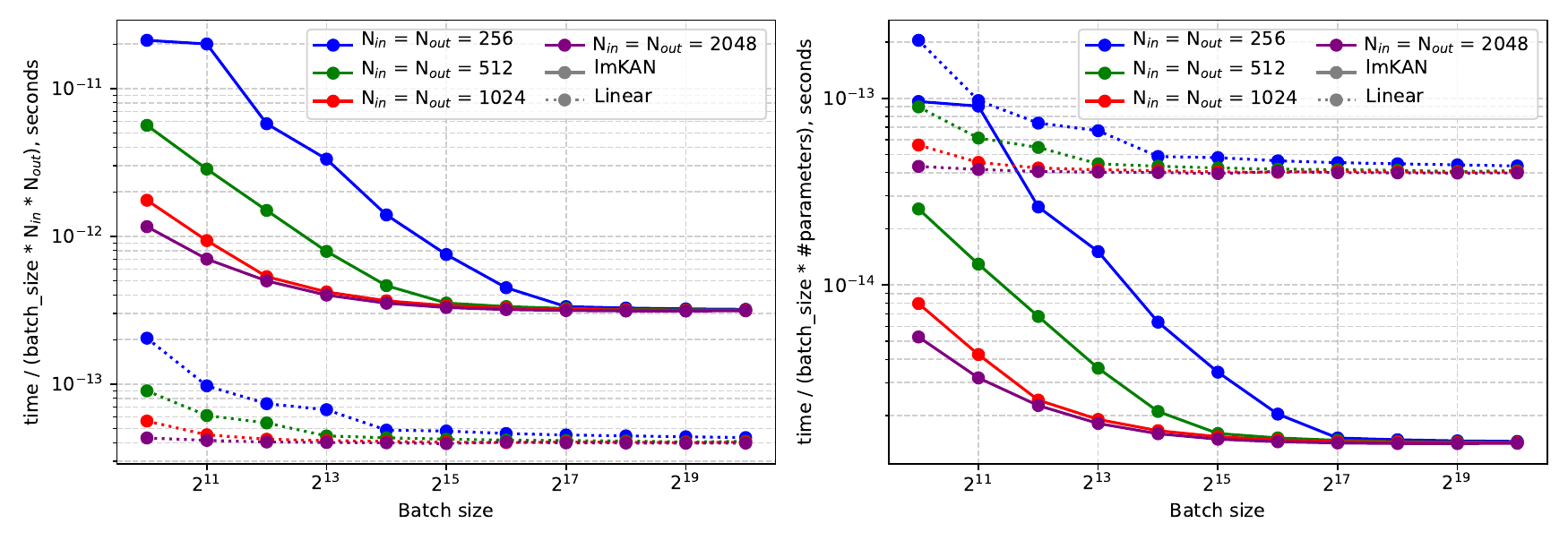}
    \caption{The performance of our CUDA kernels on the H100 SXM GPU in comparison with the linear layer in the limit of large dimensions. Left panel - time normalized by shape. Right panel - time normalized by the number of parameters.}
    \label{fig:cuda_benchmarks}
\end{figure}

The performance of our CUDA kernels with 16x16 tile size in the limit of large dimensions is summarized in Fig.~\ref{fig:cuda_benchmarks}. All the lmKAN curves are computed with the largest number of grid intervals $G = 20$ available for the 16x16 tile size. We compare the inference efficiency of lmKAN and linear layers. On the left panel, we normalize time by the shape of the layers. Fig.~\ref{fig:cuda_benchmarks} illustrates a clear convergence of these normalized times to the same value for all the dimensions. In the limit of large batch sizes, the forward pass of a lmKAN layer is \textasciitilde8x slower compared to a linear layer with the same shape. At the same time, a lmKAN layer contains a significantly larger number of parameters than a linear layer of the same shape. Thus, inference time per parameter is significantly better for lmKAN layers, about 27 times, as illustrated on the right panel of Fig.~\ref{fig:cuda_benchmarks}.

\begin{figure}[h]
    \centering
    \includegraphics[width=1.0\linewidth]{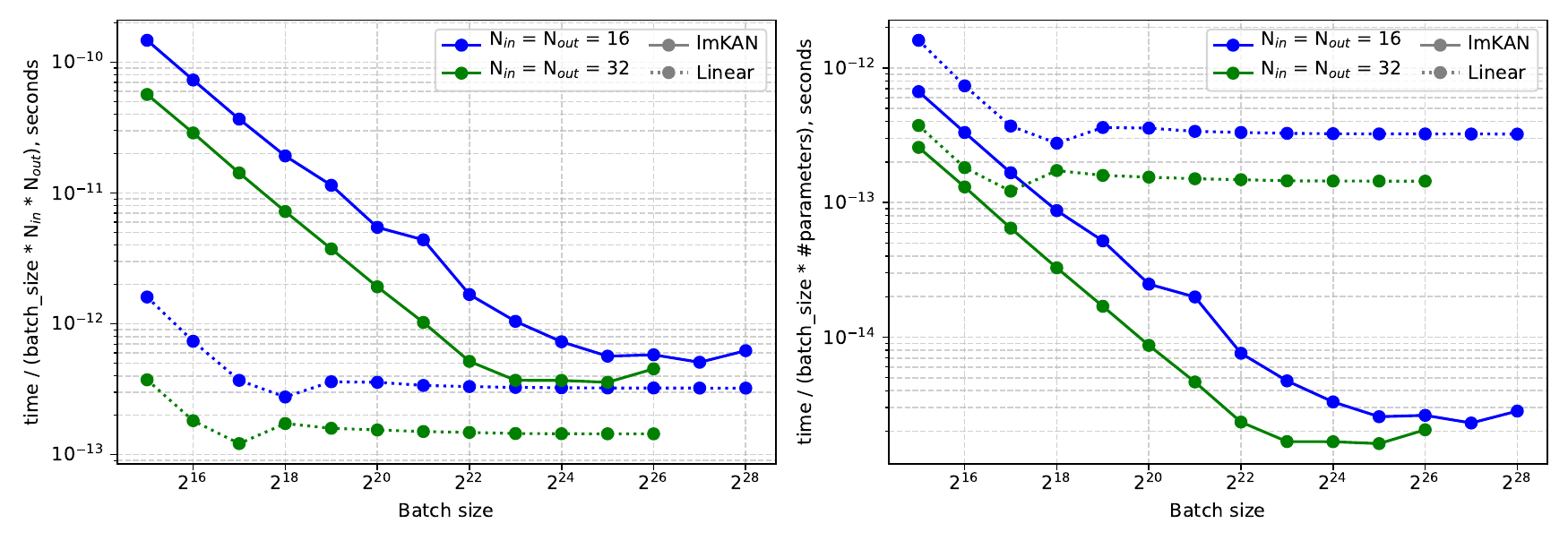}
    \caption{The performance of our CUDA kernels on the H100 SXM GPU in comparison with the linear layer for small dimensions. Left panel - time normalized by shape. Right panel - time normalized by the number of parameters.}
    \label{fig:cuda_benchmarks_small}
\end{figure}

Fig.~\ref{fig:cuda_benchmarks_small} is an analogous illustration but for small dimensions - 16 and 32. Our CUDA kernels are better adjusted for such small dimensions, and thus, relative performance compared to linear layers is even higher in this case. 

\begin{figure}[h]
    \centering
    \includegraphics[width=0.6\linewidth]{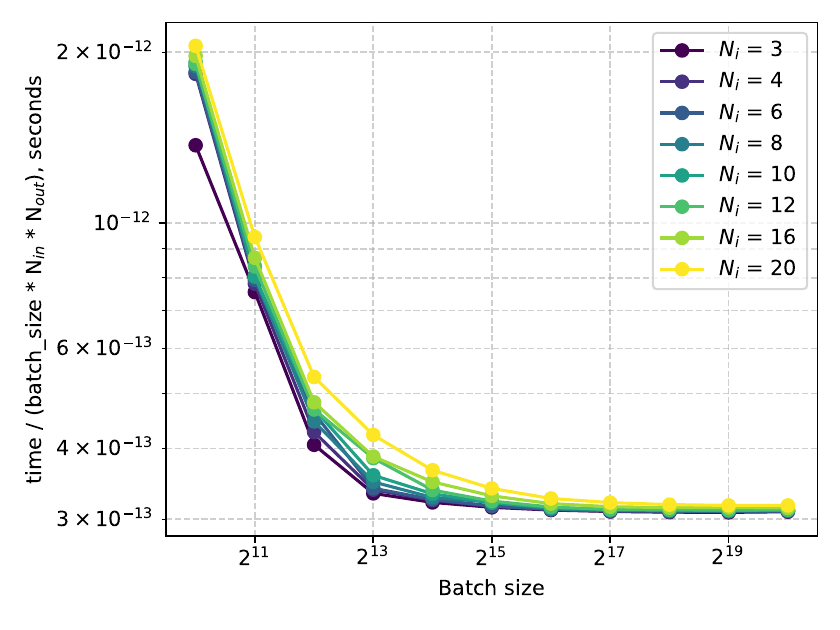}
    \caption{Inference efficiency of an lmKAN layer depending on the number of grid intervals $G$.}
    \label{fig:n_chunks_benchmark}
\end{figure}

Finally, Fig.~\ref{fig:n_chunks_benchmark} illustrates the inference efficiency depending on the number of grid intervals $G$, which control the number of parameters. The time indeed does not depend on $G$ in the large batch size limit. 

\section{Preconditioning and fitting scheme}
\label{appendix:preconditioning}

The first thing we attempted upon implementing the CUDA kernels was to fit a model with the highest grid resolution, $G=40$, supported for the $8\times8$ tile on the H100 GPU. In this setup, each 2D function had as many as $41^2=1681$ trainable parameters. We found that the training was unstable, so we designed a preconditioning and multi-stage fitting pipeline to stabilize it, which ended up being rather sophisticated. We employed this pipeline consistently for all our experiments. 

The subsequent evidence revealed that lmKANs (similarly to KANs) are progressively harder to fit as grid resolution increases. In other words, our very first experiment was the most challenging one. At more moderate grid resolutions, preconditioning measures can likely be simplified, if not omitted altogether. Specifically, we think that a fitting scheme omitting additional preconditioning terms, but preserving the Hessian regularization decay phase, which is described in the following, could be effective. With that, below is the description of the current pipeline.

\subsection{Preconditioning}

We precondition lmKAN layers by adding linear terms into the overall functional form. We use one of the following:

\begin{subequations}            
\noindent
\begin{minipage}[t]{0.45\linewidth}
  \begin{equation}
    y =\ \gamma \;\mathrm{lmKAN}(x)
          + \operatorname{ReLU}\!\bigl(\operatorname{Linear}(x)\bigr)
    \label{eq:4a}
  \end{equation}
\end{minipage}%
\hfill
\begin{minipage}[t]{0.45\linewidth}
  \begin{equation}
    y = \gamma \; \mathrm{lmKAN}(x)
          + \operatorname{Linear}\!\bigl(\operatorname{ReLU}(x)\bigr)
    \label{eq:4b}
  \end{equation}
\end{minipage}
\end{subequations}

where the lmKAN weight, $\gamma$, is initially set to $0$ and then gradually increased in our multistaged fitting procedure described later. In the case of ReLU-last preconditioning of Eq.~\ref{eq:4a}, we insert ReLU into all the layers except the last one; for ReLU-first preconditioning of Eq.~\ref{eq:4b}, we insert ReLU into all the layers except the first one. Therefore, at the beginning, when the lmKAN weight is zero, the model is equivalent to a pure MLP-based one for both types of preconditioning. 

A merit of ReLU-first preconditioning is that during inference the whole Eq.~\ref{eq:4b} can be absorbed into a single lmKAN layer whenever the number of grid intervals~$G$ is even, that is, when the origin is one of the grid points, see more details in Appendix~\ref{appendix:general_details_about_benchmarks}. Thus, this type of preconditioning does not increase inference cost in any way. This is an advantage over the original KAN preconditioning scheme~\cite{liu2024kan}, which requires the additional computation of the computationally expensive transcendental function SiLU for each edge at inference.

Because of the possibility of such an absorption, the total inference FLOPs of a ReLU-first preconditioned lmKAN layer is 2× those of a linear layer of the same shape, while for the ReLU-last preconditioning, the slowdown factor is 3×, taking into account the linear branch. 

\subsection{Fitting procedure}
\label{appendix:fitting_procedure}
Our fitting scheme consists of several phases:
\newline
\textbf{Phase I - pure MLP:} $\gamma$ is set to $0$, so the whole architecture is operating in pure MLP mode. This part is typically very short. 
\newline
\textbf{Phase II - turning on lmKAN:} $\gamma$ is linearly (over time) increased from $0$ to $0.3$. After that, there is some part with the constant $\gamma = 0.3$. At this phase, we use very strong (= with a very high coefficient $\lambda$) Hessian regularization introduced in Appendix~\ref{appendix:off_diagonal_regularization}. For all the subsequent phases, lmKAN weight $\gamma$ is fixed at $0.3$. The pipeline is also stable if $\gamma$ is increased to $1.0$, but in a few experiments we found that using $0.3$ value leads to slightly better final accuracy.
\newline
\textbf{Phase III - Hessian regularization decay:} At this phase, we gradually decay the strength of the Hessian regularization $\lambda$ from the initial very high value to the target value if this regularization is intended to be utilized in this fitting procedure and to nearly zero otherwise. 
\newline
\textbf{Phase IV - Main lmKAN fitting part:} In this final phase, we keep Hessian regularization to be constant at the value reached in the previous phase. The model is fit with the given learning rate schedule. In the experiments in this work, we use a constant learning rate for the most part of this phase, and step or exponential learning rate decay at the end. 

\begin{figure}[H]
    \centering
    \includegraphics[width=0.9\linewidth]{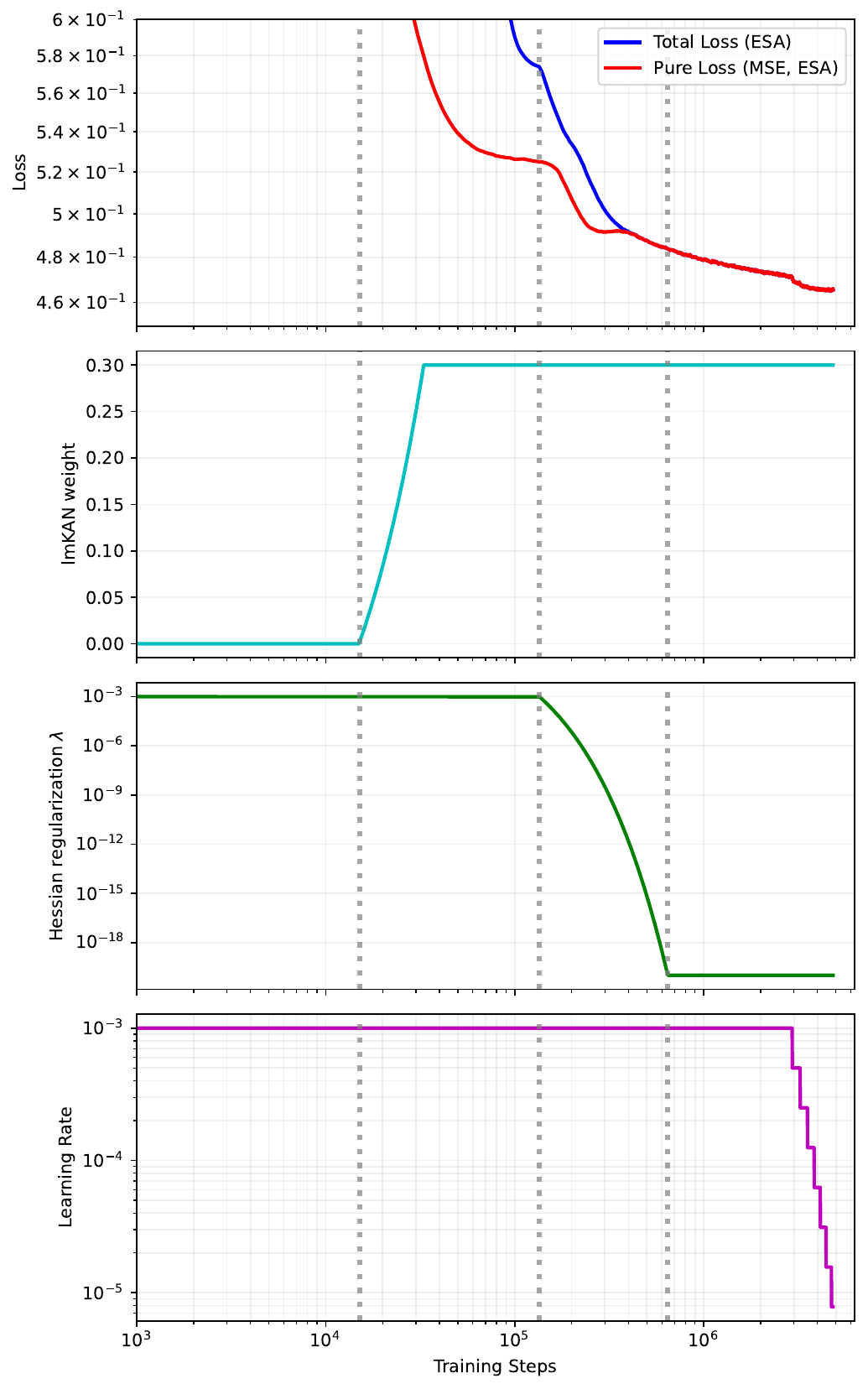}
    \caption{The multi-staged fitting procedure we use. Total loss indicates the full loss, including the Hessian regularization term. Pure loss is only the MSE part. For clarity, we plot exponential sliding averages of losses. Note that the horizontal scale is logarithmic. If it is linear, the first couple of phases are hard to discern as they are very short. The fourth phase takes most of the training budget. This training run corresponds to the unregularized lmKAN, where Hessian regularization is turned on only at the beginning of the fitting to ensure stability. At the end of phase III, it reaches nearly zero value, which, in this case, is $10^{-20}$. }
    \label{fig:fitting_scheme}
\end{figure}

An example of the described fitting procedure for one of the training runs we did for numerical experiments described in Sec.~\ref{sec:general_function_approximation} of the main text and in the Appendix~\ref{appendix:general_function_approximation} is given in Fig.~\ref{fig:fitting_scheme}. 

\subsection{Comparison of the preconditioning schemes}
\begin{figure}[H]
    \centering
    \includegraphics[width=\linewidth]{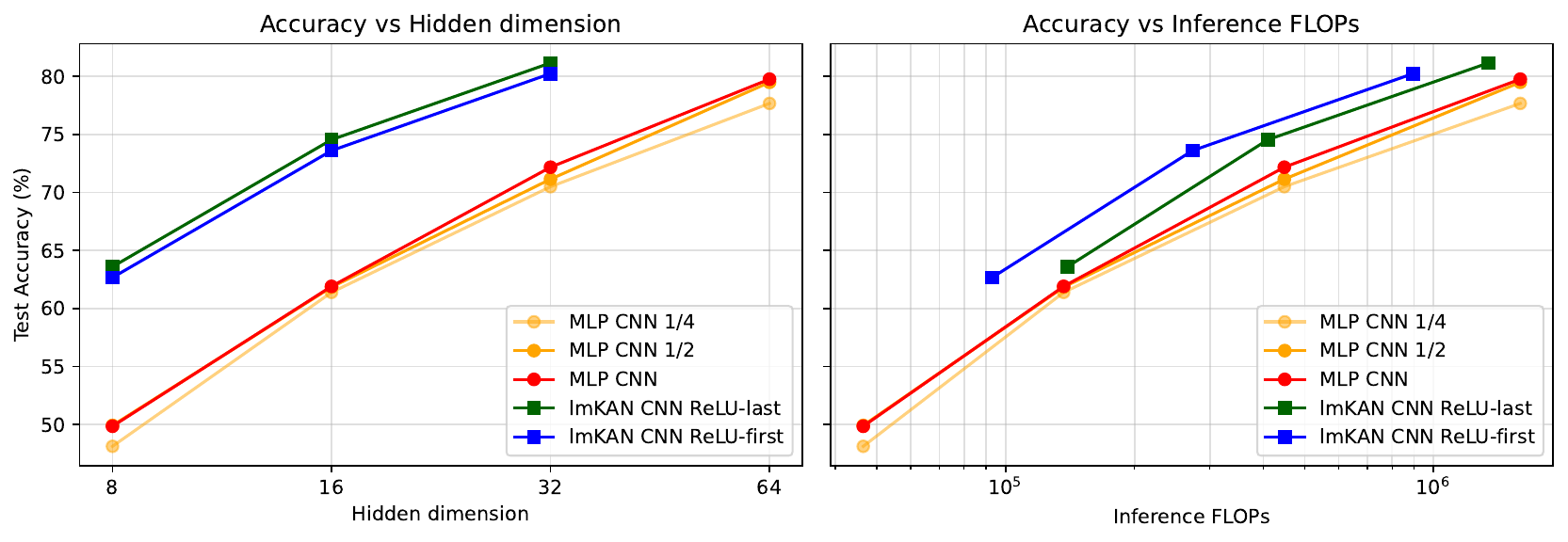}
    \caption{Comparison of the preconditioning schemes when fitting lmKANs on the CIFAR-10 dataset.}
    \label{fig:cifar10_comparison}
\end{figure}

\begin{figure}[h]
    \centering
    \includegraphics[width=\linewidth]{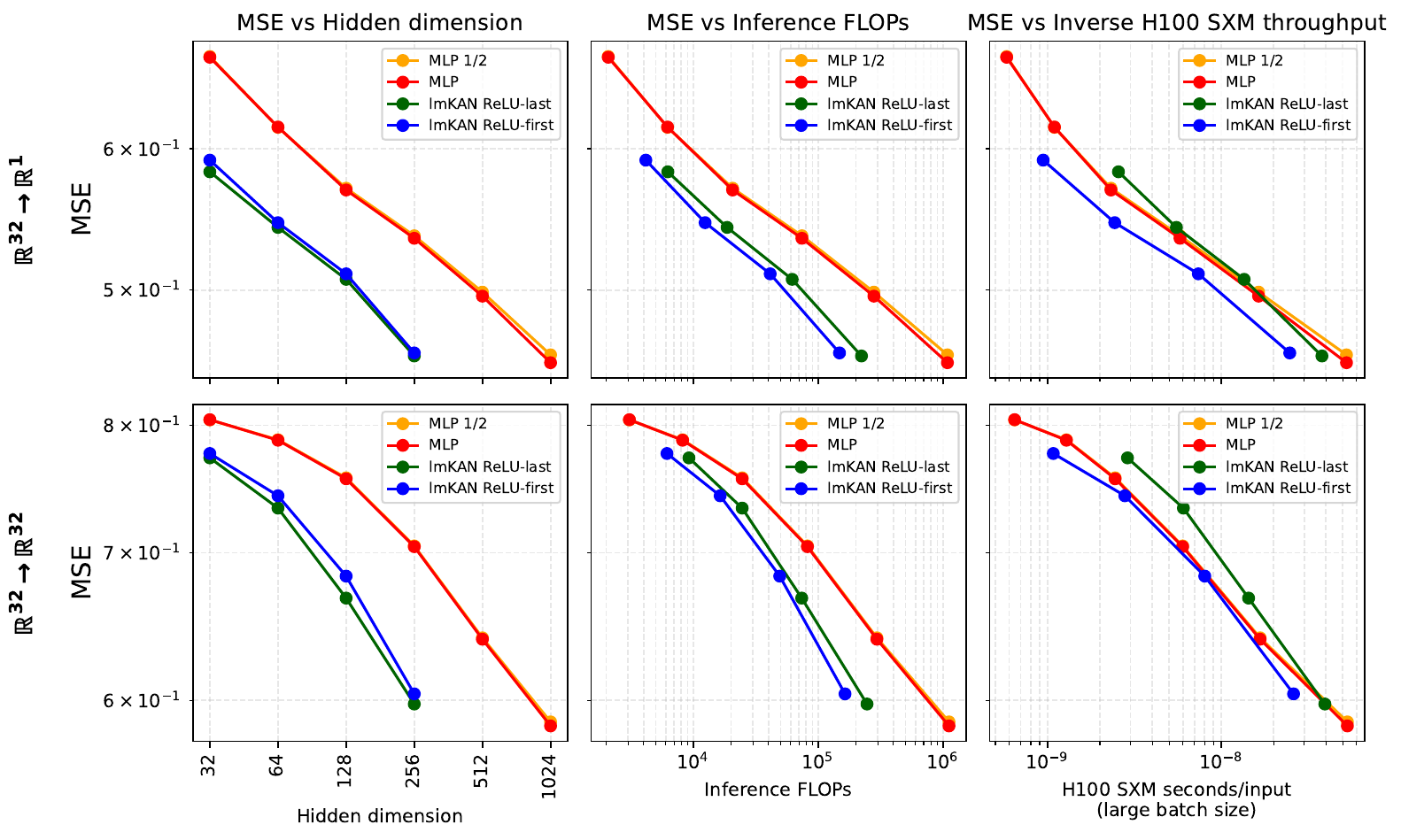}
    \caption{Comparison of the preconditioning schemes when fitting lmKANs within general function approximation setup.}
    \label{fig:mlp_teacher_comparison}
\end{figure}

We fitted lmKAN models with both types of preconditioning for the CIFAR-10 dataset and for general function approximation. The results are given in Fig.~\ref{fig:cifar10_comparison} and Fig.~\ref{fig:mlp_teacher_comparison}. 
Overall, the ReLU-last type of preconditioning appeared to lead to slightly more accurate models, but this small gain in accuracy does not justify additional computational cost. 

When designing some of our experiments we did not know this yet. Therefore, some of them use the ReLU-last type of preconditioning. 

\begin{figure}[h]
    \centering
    \includegraphics[width=0.8\linewidth]{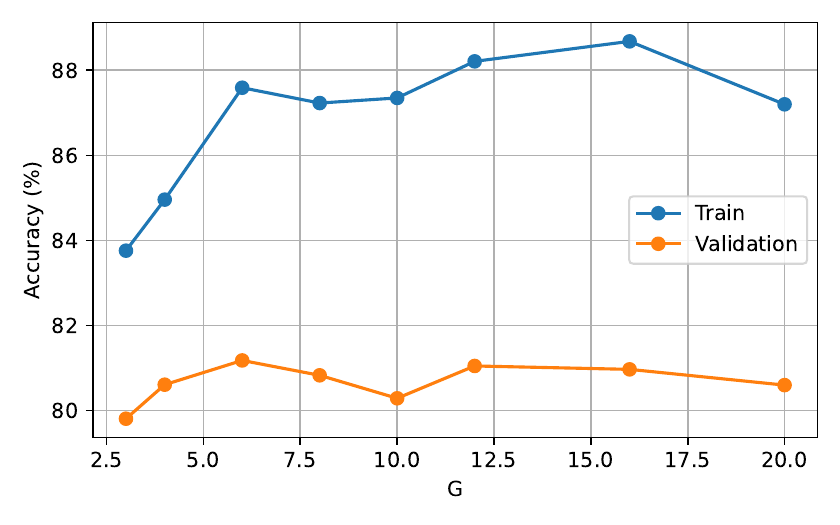}
    \caption{Accuracy(G)}
    \label{fig:cifar10_G_ablation}
\end{figure}

\begin{figure}[h]
    \centering
    \includegraphics[width=0.8\linewidth]{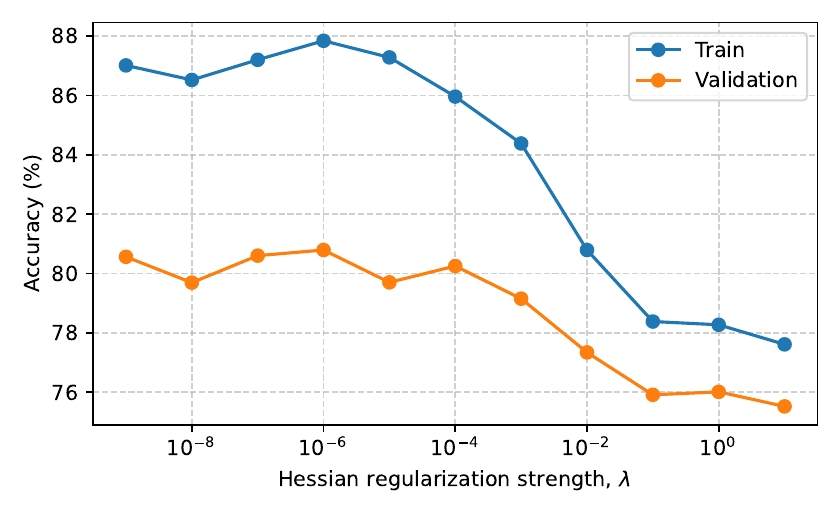}
    \caption{Accuracy(Hessian regularization strength)}
    \label{fig:cifar10_hessian_reg_ablation}
\end{figure}

We use the ReLU-last type of preconditioning for figures \ref{fig:general_function_n_chunks_plot}, \ref{fig:methane_results}, \ref{fig:methane_regularization_plot}, and \ref{fig:methane_n_chunks_grid}.
We use the ReLU-first type of preconditioning for figures \ref{fig:general_function_approximation}, \ref{fig:cnn_results}, \ref{fig:fastkan_comparison_updated}, \ref{fig:cifar10_G_ablation}, and \ref{fig:cifar10_hessian_reg_ablation}.

In other words, the performance of lmKANs on the methane datasets can likely be further improved by switching from the ReLU-last type of preconditioning to the ReLU-first one. However, since the observed gains in efficiency are already more than an order of magnitude in terms of the H100 wall-clock time, we left this for future work.

\section{Experiments}
\subsection{General details about the benchmarking protocols}
\label{appendix:general_details_about_benchmarks}
Within the scope of this work, we primarily focus on the saturated throughput in the limit of large batch sizes. Thus, we benchmark all the models for progressively large batch sizes until reaching saturation. All the models are benchmarked with \textbf{10} warm-up dry runs, and \textbf{20} timed runs. Overall, we tried to optimize each model as much as possible while staying within the limits of full precision \texttt{float32} data type. 

MLPs employed in this work consist of three types of layers - linear ones, ReLU activations, and batch normalizations. At inference, batch normalizations simply perform elementwise linear transformation and thus can be absorbed into the weights of linear layers. We perform this operation manually and, on top of that, compile the model with the \texttt{torch\_tensorrt} backend (with disabling \texttt{tf32} to ensure full precision \texttt{float32}). We use the same compilation strategy for FastKANs. 

For lmKANs, when using ReLU-first preconditioning (see more details in Appendix~\ref{appendix:preconditioning}), we absorb all the expression in Eq.~\ref{eq:4b} into the weights of the lmKAN layer. This modification requires updating the lmKAN 2D functions as $f(x_1, x_2) \leftarrow \gamma f(x_1, x_2) + w_1 \text{ReLU}(x_1) + w_2\text{ReLU}(x_2)$. Our construction allows this absorption to be done exactly whenever the origin is one of the grid points, which, in turn, is the case when the number of grid intervals $G$ is even. On top of that, batch normalizations are absorbed similarly to MLPs. We do not compile lmKAN models. 

For lmKAN models with the ReLU-last preconditioning we absorb only the lmKAN weight $\gamma$.

\subsection{General Function Approximation}
\label{appendix:general_function_approximation}

\begin{figure}[h]
    \centering
    \includegraphics[width=\linewidth]{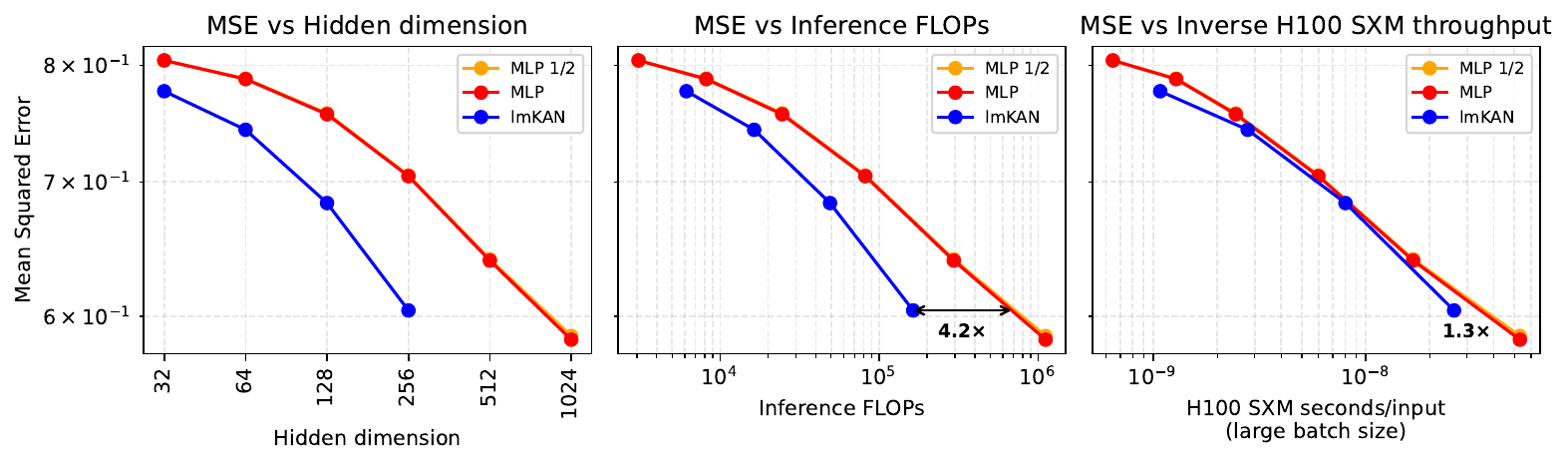}
    \caption{lmKAN vs MLP for general, $\mathbb{R}^{32} \rightarrow \mathbb{R}^{32}$, function approximation. The "MLP 1/2" line corresponds to the outcome of the fitting procedure with only half of the training steps compared to the "MLP" one.}
    \label{fig:general_function_approximation_32_32}
\end{figure}

The performance of lmKANs when approximating a $\mathbb{R}^{32} \rightarrow \mathbb{R}^{32}$ function generated similarly as the $\mathbb{R}^{32} \rightarrow \mathbb{R}^{1}$ described in the main text is given in Fig.~\ref{fig:general_function_approximation_32_32}.

There is a trend that the relative performance of lmKANs improves with the scale. It is clearly seen on the MSE vs FLOPs panel. On the MSE vs H100 wall-clock time panel, it is first masked by the non-homogeneous efficiency of the code, but next still reveals itself for the largest hidden dimensions. 

\subsection{Methane}
\label{appendix:methane}

\subsubsection{The \texttt{Distances Polynomials} representation}
It was mentioned in Sec.~\ref {sec:methane} of the main text that the representation {\ttfamily\bfseries Distances Polynomials} is given by non-trivial invariant polynomials computed on top of interatomic distances. These polynomials are constant with respect to changing the order of identical hydrogen atoms. 

\begin{equation}
\label{eq:polynomials_1}
\begin{aligned}
P_{1} &= x_{5} + x_{6} + x_{7} + x_{8} + x_{9} + x_{10} \\
P_{2} &= x_{1} + x_{2} + x_{3} + x_{4} \\
P_{3} &= x_{5}^{2} + x_{6}^{2} + x_{7}^{2} + x_{8}^{2} + x_{9}^{2} + x_{10}^{2} \\
P_{4} &= x_{5} x_{6} + x_{5} x_{7} + x_{6} x_{7} + x_{5} x_{8} + x_{6} x_{8} + x_{5} x_{9} + x_{7} x_{9} + x_{8} x_{9} + \\
 &\phantom{=} x_{6} x_{10} + x_{7} x_{10} + x_{8} x_{10} + x_{9} x_{10} \\
P_{5} &= x_{1} x_{5} + x_{2} x_{5} + x_{1} x_{6} + x_{3} x_{6} + x_{1} x_{7} + x_{4} x_{7} + x_{2} x_{8} + x_{3} x_{8} + \\
 &\phantom{=} x_{2} x_{9} + x_{4} x_{9} + x_{3} x_{10} + x_{4} x_{10} \\
P_{6} &= x_{1}^{2} + x_{2}^{2} + x_{3}^{2} + x_{4}^{2} \\
P_{7} &= x_{5}^{3} + x_{6}^{3} + x_{7}^{3} + x_{8}^{3} + x_{9}^{3} + x_{10}^{3} \\
P_{8} &= x_{5}^{2} x_{6} + x_{5} x_{6}^{2} + x_{5}^{2} x_{7} + x_{6}^{2} x_{7} + x_{5} x_{7}^{2} + x_{6} x_{7}^{2} + x_{5}^{2} x_{8} + x_{6}^{2} x_{8} + \\
 &\phantom{=} x_{5} x_{8}^{2} + x_{6} x_{8}^{2} + x_{5}^{2} x_{9} + x_{7}^{2} x_{9} + x_{8}^{2} x_{9} + x_{5} x_{9}^{2} + x_{7} x_{9}^{2} + x_{8} x_{9}^{2} + \\
 &\phantom{=} x_{6}^{2} x_{10} + x_{7}^{2} x_{10} + x_{8}^{2} x_{10} + x_{9}^{2} x_{10} + x_{6} x_{10}^{2} + x_{7} x_{10}^{2} + x_{8} x_{10}^{2} + x_{9} x_{10}^{2} \\
P_{9} &= x_{1} x_{5}^{2} + x_{2} x_{5}^{2} + x_{1} x_{6}^{2} + x_{3} x_{6}^{2} + x_{1} x_{7}^{2} + x_{4} x_{7}^{2} + x_{2} x_{8}^{2} + x_{3} x_{8}^{2} + \\
 &\phantom{=} x_{2} x_{9}^{2} + x_{4} x_{9}^{2} + x_{3} x_{10}^{2} + x_{4} x_{10}^{2} \\
P_{10} &= x_{5} x_{6} x_{8} + x_{5} x_{7} x_{9} + x_{6} x_{7} x_{10} + x_{8} x_{9} x_{10} \\
P_{11} &= x_{1} x_{5} x_{6} + x_{1} x_{5} x_{7} + x_{1} x_{6} x_{7} + x_{2} x_{5} x_{8} + x_{3} x_{6} x_{8} + x_{2} x_{5} x_{9} + x_{4} x_{7} x_{9} + x_{2} x_{8} x_{9} + \\
 &\phantom{=} x_{3} x_{6} x_{10} + x_{4} x_{7} x_{10} + x_{3} x_{8} x_{10} + x_{4} x_{9} x_{10} \\
P_{12} &= x_{1}^{2} x_{5} + x_{2}^{2} x_{5} + x_{1}^{2} x_{6} + x_{3}^{2} x_{6} + x_{1}^{2} x_{7} + x_{4}^{2} x_{7} + x_{2}^{2} x_{8} + x_{3}^{2} x_{8} + \\
 &\phantom{=} x_{2}^{2} x_{9} + x_{4}^{2} x_{9} + x_{3}^{2} x_{10} + x_{4}^{2} x_{10} \\
P_{13} &= x_{1} x_{2} x_{5} + x_{1} x_{3} x_{6} + x_{1} x_{4} x_{7} + x_{2} x_{3} x_{8} + x_{2} x_{4} x_{9} + x_{3} x_{4} x_{10} \\
P_{14} &= x_{1}^{3} + x_{2}^{3} + x_{3}^{3} + x_{4}^{3} \\
P_{15} &= x_{5}^{4} + x_{6}^{4} + x_{7}^{4} + x_{8}^{4} + x_{9}^{4} + x_{10}^{4} \\
P_{16} &= x_{5}^{3} x_{6} + x_{5} x_{6}^{3} + x_{5}^{3} x_{7} + x_{6}^{3} x_{7} + x_{5} x_{7}^{3} + x_{6} x_{7}^{3} + x_{5}^{3} x_{8} + x_{6}^{3} x_{8} + \\
 &\phantom{=} x_{5} x_{8}^{3} + x_{6} x_{8}^{3} + x_{5}^{3} x_{9} + x_{7}^{3} x_{9} + x_{8}^{3} x_{9} + x_{5} x_{9}^{3} + x_{7} x_{9}^{3} + x_{8} x_{9}^{3} + \\
 &\phantom{=} x_{6}^{3} x_{10} + x_{7}^{3} x_{10} + x_{8}^{3} x_{10} + x_{9}^{3} x_{10} + x_{6} x_{10}^{3} + x_{7} x_{10}^{3} + x_{8} x_{10}^{3} + x_{9} x_{10}^{3} \\
P_{17} &= x_{1} x_{5}^{3} + x_{2} x_{5}^{3} + x_{1} x_{6}^{3} + x_{3} x_{6}^{3} + x_{1} x_{7}^{3} + x_{4} x_{7}^{3} + x_{2} x_{8}^{3} + x_{3} x_{8}^{3} + \\
 &\phantom{=} x_{2} x_{9}^{3} + x_{4} x_{9}^{3} + x_{3} x_{10}^{3} + x_{4} x_{10}^{3} \\
P_{18} &= x_{1} x_{5}^{2} x_{6} + x_{1} x_{5} x_{6}^{2} + x_{1} x_{5}^{2} x_{7} + x_{1} x_{6}^{2} x_{7} + x_{1} x_{5} x_{7}^{2} + x_{1} x_{6} x_{7}^{2} + x_{2} x_{5}^{2} x_{8} + x_{3} x_{6}^{2} x_{8} + \\
 &\phantom{=} x_{2} x_{5} x_{8}^{2} + x_{3} x_{6} x_{8}^{2} + x_{2} x_{5}^{2} x_{9} + x_{4} x_{7}^{2} x_{9} + x_{2} x_{8}^{2} x_{9} + x_{2} x_{5} x_{9}^{2} + x_{4} x_{7} x_{9}^{2} + x_{2} x_{8} x_{9}^{2} + \\
 &\phantom{=} x_{3} x_{6}^{2} x_{10} + x_{4} x_{7}^{2} x_{10} + x_{3} x_{8}^{2} x_{10} + x_{4} x_{9}^{2} x_{10} + x_{3} x_{6} x_{10}^{2} + x_{4} x_{7} x_{10}^{2} + x_{3} x_{8} x_{10}^{2} + x_{4} x_{9} x_{10}^{2} \\
P_{19} &= x_{2} x_{5}^{2} x_{6} + x_{3} x_{5} x_{6}^{2} + x_{2} x_{5}^{2} x_{7} + x_{3} x_{6}^{2} x_{7} + x_{4} x_{5} x_{7}^{2} + x_{4} x_{6} x_{7}^{2} + x_{1} x_{5}^{2} x_{8} + x_{1} x_{6}^{2} x_{8} + \\
 &\phantom{=} x_{3} x_{5} x_{8}^{2} + x_{2} x_{6} x_{8}^{2} + x_{1} x_{5}^{2} x_{9} + x_{1} x_{7}^{2} x_{9} + x_{3} x_{8}^{2} x_{9} + x_{4} x_{5} x_{9}^{2} + x_{2} x_{7} x_{9}^{2} + x_{4} x_{8} x_{9}^{2} + \\
 &\phantom{=} x_{1} x_{6}^{2} x_{10} + x_{1} x_{7}^{2} x_{10} + x_{2} x_{8}^{2} x_{10} + x_{2} x_{9}^{2} x_{10} + x_{4} x_{6} x_{10}^{2} + x_{3} x_{7} x_{10}^{2} + x_{4} x_{8} x_{10}^{2} + x_{3} x_{9} x_{10}^{2} \\
P_{20} &= x_{1}^{2} x_{5}^{2} + x_{2}^{2} x_{5}^{2} + x_{1}^{2} x_{6}^{2} + x_{3}^{2} x_{6}^{2} + x_{1}^{2} x_{7}^{2} + x_{4}^{2} x_{7}^{2} + x_{2}^{2} x_{8}^{2} + x_{3}^{2} x_{8}^{2} + \\
 &\phantom{=} x_{2}^{2} x_{9}^{2} + x_{4}^{2} x_{9}^{2} + x_{3}^{2} x_{10}^{2} + x_{4}^{2} x_{10}^{2}
\end{aligned}
\end{equation}

\begin{equation}
\label{eq:polynomials_2}
\begin{aligned}
P_{21} &= x_{1} x_{2} x_{5}^{2} + x_{1} x_{3} x_{6}^{2} + x_{1} x_{4} x_{7}^{2} + x_{2} x_{3} x_{8}^{2} + x_{2} x_{4} x_{9}^{2} + x_{3} x_{4} x_{10}^{2} \\
P_{22} &= x_{1}^{2} x_{5} x_{6} + x_{1}^{2} x_{5} x_{7} + x_{1}^{2} x_{6} x_{7} + x_{2}^{2} x_{5} x_{8} + x_{3}^{2} x_{6} x_{8} + x_{2}^{2} x_{5} x_{9} + x_{4}^{2} x_{7} x_{9} + x_{2}^{2} x_{8} x_{9} + \\
 &\phantom{=} x_{3}^{2} x_{6} x_{10} + x_{4}^{2} x_{7} x_{10} + x_{3}^{2} x_{8} x_{10} + x_{4}^{2} x_{9} x_{10} \\
P_{23} &= x_{1}^{3} x_{5} + x_{2}^{3} x_{5} + x_{1}^{3} x_{6} + x_{3}^{3} x_{6} + x_{1}^{3} x_{7} + x_{4}^{3} x_{7} + x_{2}^{3} x_{8} + x_{3}^{3} x_{8} + \\
 &\phantom{=} x_{2}^{3} x_{9} + x_{4}^{3} x_{9} + x_{3}^{3} x_{10} + x_{4}^{3} x_{10} \\
P_{24} &= x_{1}^{4} + x_{2}^{4} + x_{3}^{4} + x_{4}^{4} \\
P_{25} &= x_{5}^{5} + x_{6}^{5} + x_{7}^{5} + x_{8}^{5} + x_{9}^{5} + x_{10}^{5} \\
P_{26} &= x_{1} x_{5}^{4} + x_{2} x_{5}^{4} + x_{1} x_{6}^{4} + x_{3} x_{6}^{4} + x_{1} x_{7}^{4} + x_{4} x_{7}^{4} + x_{2} x_{8}^{4} + x_{3} x_{8}^{4} + \\
 &\phantom{=} x_{2} x_{9}^{4} + x_{4} x_{9}^{4} + x_{3} x_{10}^{4} + x_{4} x_{10}^{4} \\
P_{27} &= x_{1} x_{5}^{3} x_{6} + x_{1} x_{5} x_{6}^{3} + x_{1} x_{5}^{3} x_{7} + x_{1} x_{6}^{3} x_{7} + x_{1} x_{5} x_{7}^{3} + x_{1} x_{6} x_{7}^{3} + x_{2} x_{5}^{3} x_{8} + x_{3} x_{6}^{3} x_{8} + \\
 &\phantom{=} x_{2} x_{5} x_{8}^{3} + x_{3} x_{6} x_{8}^{3} + x_{2} x_{5}^{3} x_{9} + x_{4} x_{7}^{3} x_{9} + x_{2} x_{8}^{3} x_{9} + x_{2} x_{5} x_{9}^{3} + x_{4} x_{7} x_{9}^{3} + x_{2} x_{8} x_{9}^{3} + \\
 &\phantom{=} x_{3} x_{6}^{3} x_{10} + x_{4} x_{7}^{3} x_{10} + x_{3} x_{8}^{3} x_{10} + x_{4} x_{9}^{3} x_{10} + x_{3} x_{6} x_{10}^{3} + x_{4} x_{7} x_{10}^{3} + x_{3} x_{8} x_{10}^{3} + x_{4} x_{9} x_{10}^{3} \\
P_{28} &= x_{1}^{2} x_{5}^{3} + x_{2}^{2} x_{5}^{3} + x_{1}^{2} x_{6}^{3} + x_{3}^{2} x_{6}^{3} + x_{1}^{2} x_{7}^{3} + x_{4}^{2} x_{7}^{3} + x_{2}^{2} x_{8}^{3} + x_{3}^{2} x_{8}^{3} + \\
 &\phantom{=} x_{2}^{2} x_{9}^{3} + x_{4}^{2} x_{9}^{3} + x_{3}^{2} x_{10}^{3} + x_{4}^{2} x_{10}^{3} \\
P_{29} &= x_{1} x_{2} x_{5}^{3} + x_{1} x_{3} x_{6}^{3} + x_{1} x_{4} x_{7}^{3} + x_{2} x_{3} x_{8}^{3} + x_{2} x_{4} x_{9}^{3} + x_{3} x_{4} x_{10}^{3} \\
P_{30} &= x_{1}^{3} x_{5}^{2} + x_{2}^{3} x_{5}^{2} + x_{1}^{3} x_{6}^{2} + x_{3}^{3} x_{6}^{2} + x_{1}^{3} x_{7}^{2} + x_{4}^{3} x_{7}^{2} + x_{2}^{3} x_{8}^{2} + x_{3}^{3} x_{8}^{2} + \\
 &\phantom{=} x_{2}^{3} x_{9}^{2} + x_{4}^{3} x_{9}^{2} + x_{3}^{3} x_{10}^{2} + x_{4}^{3} x_{10}^{2} \\
P_{31} &= x_{1}^{3} x_{2} x_{5} + x_{1} x_{2}^{3} x_{5} + x_{1}^{3} x_{3} x_{6} + x_{1} x_{3}^{3} x_{6} + x_{1}^{3} x_{4} x_{7} + x_{1} x_{4}^{3} x_{7} + x_{2}^{3} x_{3} x_{8} + x_{2} x_{3}^{3} x_{8} + \\
 &\phantom{=} x_{2}^{3} x_{4} x_{9} + x_{2} x_{4}^{3} x_{9} + x_{3}^{3} x_{4} x_{10} + x_{3} x_{4}^{3} x_{10}
\end{aligned}
\end{equation}

The exact form of these polynomials is given in Eq.~\ref{eq:polynomials_1} and Eq.~\ref{eq:polynomials_2}, where $x_1$, $x_2$, ... $x_{10}$ correspond to interatomic distances $CH_1$, $CH_2$, $CH_3$, $CH_4$, $H_1H_2$, $H_1H_3$, $H_1H_4$, $H_2H_3$, $H_2H_4$, and $H_3H_4$ respectively. The presented polynomials form invariant generators~\cite{derksen2015computational} of the group corresponding to arbitrary permutations of the hydrogen atoms. Therefore, the {\ttfamily\bfseries Distances Polynomials} representation preserves all the information about the initial $CH_4$ molecule. 

\subsubsection{Ablations}
\begin{figure}[h]
    \centering
    \includegraphics[width=0.6\linewidth]{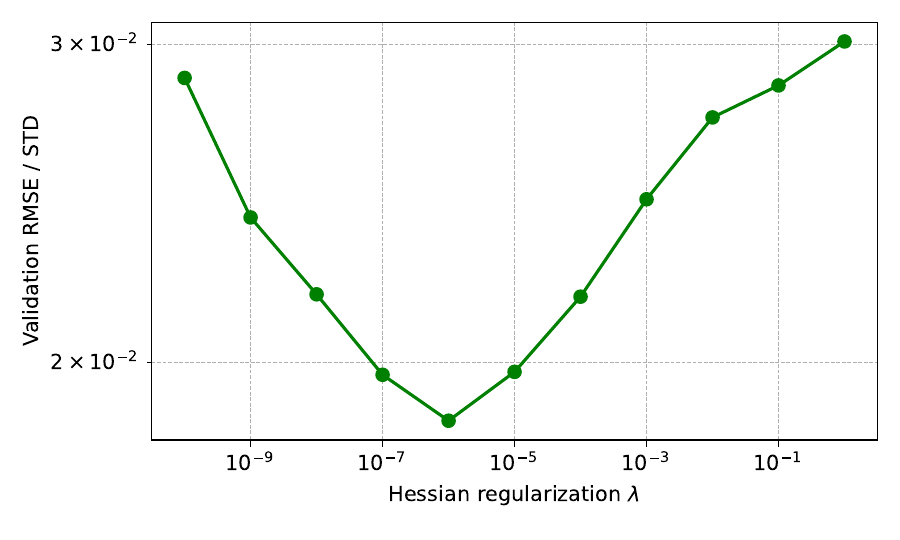}
    \caption{Effect of the strength of Hessian regularization on the validation error when fitting lmKAN with hidden\_dim = 256 on the methane dataset using the {\ttfamily\bfseries Distances Polynomials} representation.}
    \label{fig:methane_regularization_plot}
\end{figure}

\begin{figure}[h]
    \centering
    \includegraphics[width=0.6\linewidth]{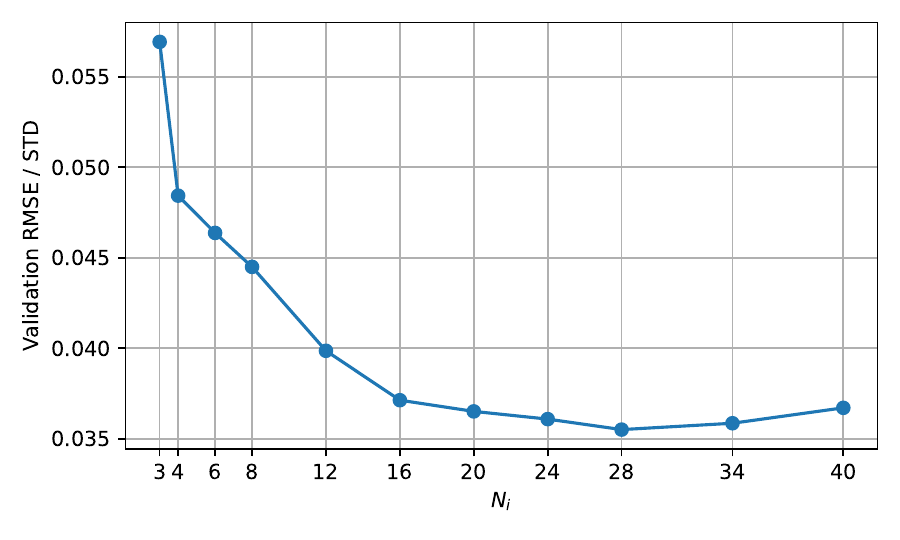}
    \caption{Effect of the number of grid intervals $G$ on the validation error when fitting lmKAN with hidden\_dim = 128 on the methane dataset using the Cartesian Components representation.}
    \label{fig:methane_n_chunks_grid}
\end{figure}

The ablation study on the effect of Hessian regularization is given in Fig.~\ref{fig:methane_regularization_plot}, and the effect of the number of grid intervals $G$ in Fig.~\ref{fig:methane_n_chunks_grid}. 

\subsection{CIFAR-10}
\label{appendix:cifar_10}

\paragraph{Augmentations}
We use the following pool of augmentations for the CIFAR-10 dataset: 

\begin{archboxaug}{CIFAR-10 augmentation pipeline}
MEAN = (0.4914, 0.4822, 0.4465)
STD = (0.2470, 0.2435, 0.2616)

nn.Sequential(
    T.RandomCrop(32, padding=4),
    T.RandomHorizontalFlip(),
    T.ColorJitter(0.3, 0.3, 0.3, 0.05),
    T.RandAugment(2, 7),
    T.RandomErasing(p=0.25, scale=(0.05, 0.2), ratio=(0.3, 3.3)),
    T.Normalize(MEAN, STD),
)
\end{archboxaug}

On top of these, we use MixUp~\cite{zhang2017mixup} ($\alpha = 0.2$) and CutMix~\cite{yun2019cutmix} ($\beta = 1.0$) augmentations, both with 50\% probability.

When fitting the families of convolutional neural networks described in the main text, we use the above pool of augmentations consistently for MLP-based and lmKAN-based CNNs. For all the details about the fitting procedures see the configuration files attached to these appendices. 

\subsection{ImageNet}
\label{appendix:imagenet}
The standard data preparation pipeline introduced by AlexNet\cite{krizhevsky2012imagenet} involves first resizing an image to $256$ pixels along the smallest dimension, then performing a random crop of $224\times224$ pixels during training, and a center crop of $224\times224$ pixels during validation. 

We mimic this procedure by first resizing the image to $81*256/224\approx93$ pixels across the smallest dimension, and then performing random or center crops of $81\times81$ pixels. 

Next, we use the following augmentation pipeline:

\begin{archboxaug}{ImageNet augmentation pipeline}
nn.Sequential(
    T.RandomHorizontalFlip(),
    T.ColorJitter(brightness=0.4, contrast=0.4, saturation=0.4, hue=0.1),
    T.RandAugment(),
    T.RandomErasing(p=0.25, scale=(0.02, 0.33), ratio=(0.3, 3.3), value=0),
    T.Normalize(mean=[0.485, 0.456, 0.406], std=[0.229, 0.224, 0.225]),
)
\end{archboxaug}

On top of these, we use MixUp and CutMix.  
\subsection{Comparison with FastKAN}
\label{appendix:fastkan}
The data was split properly into train, validation, and test subsets, while the original script employed only a train-validation split. We use the cosine (without restarts) learning rate scheduler~\cite{loshchilov2016sgdr} instead of the exponential decay of the original script. Finally, the normalization was performed with true values of the mean and standard deviation for the CIFAR-10 dataset, instead of the placeholder $0.5$ values of the original script. 

\end{document}